\documentclass[sigconf]{acmart}

\AtBeginDocument{%
  }

\setcopyright{acmlicensed}

\copyrightyear{2026}
\acmYear{2026}
\setcopyright{cc}
\setcctype{by-nc-nd}

\acmDOI{10.1145/3767308.3834743}
\acmConference[MM '26]{Proceedings of the 34th ACM International Conference on Multimedia}{November 10--14, 2026}{Rio de Janeiro, Brazil}
\acmBooktitle{Proceedings of the 34th ACM International Conference on Multimedia (MM '26), November 10--14, 2026, Rio de Janeiro, Brazil}
\acmDOI{10.1145/3767308.3834743}
\acmISBN{979-8-4007-2213-4/2026/11}

\usepackage[table]{xcolor}
\usepackage{pifont}
\usepackage{multirow}

\definecolor{forestgreen}{RGB}{34,139,34}
\definecolor{yechanmandarin}{RGB}{255,195,27}
\definecolor{yechanblue}{RGB}{23,127,254}
\definecolor{amod}{RGB}{77,77,77}
\definecolor{amodstar}{RGB}{0,191,174}
\newcommand{\symbox}[1]{\makebox[1.2em][c]{#1}} 
\newcommand{\cmark}{\textcolor{forestgreen}{\symbox{\ding{51}}}}  % ✓
\newcommand{\xmark}{\textcolor{red}{\symbox{\ding{55}}}}  % ✗

\usepackage{pifont}
\usepackage{multirow}

\begin{document}

%%
%% The "title" command has an optional parameter,
%% allowing the author to define a "short title" to be used in page headers.
\title{G-MAD: A Game-Based Data Generation Framework \\for Multi-View RGB-T Aerial Object Detection}

% \author{Submitted to ACM MM - Open Source Software Track}

% \author{Yechan Kim}\authornotemark[2]\orcid{0000-0002-2438-3590}
% \author{JongHyun Park}\authornotemark[3]\orcid{0009-0005-5404-0707}

% % \authornote{Both authors contributed equally to this research.}
% % \author{G.K.M. Tobin}
% \affiliation{%
%   \institution{$^{\dagger}$LIG Defense\&Aerospace}
%   % \country{Republic of Korea}
% }
% \affiliation{%
%   \institution{$^{\ddagger}$GIST}
%   \country{Republic of Korea}
% }

% \author{Dongho Yoon}\authornotemark[3]\orcid{0009-0006-1514-7293}
% \author{Namhoon Jung}\authornotemark[2]
% \author{Moongu Jeon}\authornotemark[3]\orcid{0000-0002-2775-7789}

% \email{{yechan.kim26, namhoon.jung}@ligdna.com}
% \email{{citizen135, gidong76}@gm.gist.ac.kr, mgjeon@gist.ac.kr}

%%
%% The "author" command and its associated commands are used to define
%% the authors and their affiliations.
%% Of note is the shared affiliation of the first two authors, and the
%% "authornote" and "authornotemark" commands
%% used to denote shared contribution to the research.

\author{Yechan Kim}
% \authornote{Both authors contributed equally to this research.}
\orcid{0000-0002-2438-3590}
% \author{G.K.M. Tobin}
% \authornotemark[1]
\affiliation{%
  \institution{LIG Defense\&Aerospace}
  \country{Republic of Korea}
}
\email{yechan.kim26@ligdna.com}

\author{JongHyun Park}
\orcid{0009-0005-5404-0707}
\affiliation{%
  \institution{Gwangju Institute of Science and Technology}
  \country{Republic of Korea}}
\email{citizen135@gm.gist.ac.kr}

\author{Dongho Yoon}
\orcid{0009-0006-1514-7293}
\affiliation{%
  \institution{Gwangju Institute of Science and Technology}
  \country{Republic of Korea}}
\email{gidong76@gm.gist.ac.kr}

\author{Namhoon Jung}
% https://orcid.org/0000-0003-1629-2418
\orcid{https://orcid.org/0000-0003-1629-2418}
\affiliation{%
  \institution{LIG Defense\&Aerospace}
  \country{Republic of Korea}
}
\email{namhoon.jung@ligdna.com}

\author{Moongu Jeon}
\orcid{0000-0002-2775-7789}
\affiliation{%
  \institution{Gwangju Institute of Science and Technology}
  \country{Republic of Korea}}
\email{mgjeon@gist.ac.kr}

%%
%% By default, the full list of authors will be used in the page
%% headers. Often, this list is too long, and will overlap
%% other information printed in the page headers. This command allows
%% the author to define a more concise list
%% of authors' names for this purpose.
\renewcommand{\shortauthors}{Yechan Kim, JongHyun Park, Dongho Yoon, Namhoon Jung, and Moongu Jeon}

%%
%% The abstract is a short summary of the work to be presented in the
%% article.
\begin{abstract}
This work introduces G-MAD, an open-source framework that uses Arma3 to generate synchronized multi-view RGB-T data for aerial object detection. G-MAD addresses key limitations of real-world aerial dataset construction, including limited viewpoint control, imperfect RGB-T alignment and high annotation cost. The framework supports structured scenario specification, controllable multi-view camera placement, simultaneous visible/thermal capture, and automatic bounding box annotation using engine-level geometric metadata. These capabilities enable controlled studies of viewpoint variation, multi-modal fusion, and synthetic-to-real transfer in aerial object detection. Besides, using G-MAD, we construct and release AMOD, a new large-scale multi-view aerial RGB-T object detection benchmark. The source code and the dataset are available at \url{https://unique-chan.github.io/G-MAD-Project}.

% \url{https://anonymous.4open.science/r/G-MAD-245E}.

% Progress in military perception is often constrained not only by model design, but by who can access the data. Real military imagery is costly, sensitive, and operationally restricted, leaving many researchers without the means to study data-centric questions at scale, particularly in multi-modal settings such as RGB and thermal-infrared perception. We present G-MAD, an open-source software framework for generating controllable synthetic RGB-T military object detection datasets with automatic annotations using a game Arma3. G-MAD enables users to flexibly compose military assets and environments, configure multi-view captures, and scalably generate datasets with diverse geometric, contextual, and cross-modal variation. By transforming restricted data collection problems into reproducible simulation workflows, the framework broadens participation in military vision research beyond highly specialized organizations. To further demonstrate the usefulness of G-MAD, we leverage it to construct and release AMOD, a new aerial-view RGB-T military object detection benchmark. We believe G-MAD and AMOD provide a practical step toward democratizing synthetic-data-driven research for RGB-T military perception systems. The source code and the dataset will be available through our official website (GitHub / HuggingFace) upon acceptance.

% are available at \url{https://unique-chan.github.io/G-MAD-Project}.
\end{abstract} 

%%
%% The code below is generated by the tool at http://dl.acm.org/ccs.cfm.
%% Please copy and paste the code instead of the example below.
%%


%%
%% Keywords. The author(s) should pick words that accurately describe
%% the work being presented. Separate the keywords with commas.

%% A "teaser" image appears between the author and affiliation
%% information and the body of the document, and typically spans the
%% page.

% \received{1 April 2026}
% \received[revised]{12 March 2009}
% \received[accepted]{5 June 2009}

%%
%% This command processes the author and affiliation and title
%% information and builds the first part of the formatted document.
\maketitle

% \vspace{-0.2cm}

\section{Introduction} \vspace{-0.0cm}

Aerial object detection has become an important problem in remote sensing, surveillance, disaster response, and autonomous driving.
    Compared with ground-level perception, aerial imagery involves large viewpoint changes, scale variation, and complex backgrounds, making robust object detection challenging.
These challenges become more pronounced in RGB-thermal (RGB-T) settings, where models must exploit complementary visible and thermal cues while maintaining accurate cross-modal correspondence.
    However, constructing real-world aerial RGB-T datasets is costly and difficult because it requires synchronized sensor acquisition, repeated flights under controlled conditions, and labor-intensive annotation. 
As a result, existing datasets often provide limited control over viewpoints, environmental factors, and modality alignment.

Synthetic data generation provides a practical way to address these limitations \cite{paulin2023review, delussu2024synthetic}. 
    While recent approaches span a broad spectrum---from classical 3D graphics and simulation pipelines to neural rendering and generative models---not all methods provide the structured supervision and controllability required for perception tasks. 
In particular, simulation-based pipelines remain well suited for such settings, as they provide direct access to scene geometry, annotations, and sensor configurations \cite{dosovitskiy2017carla}.

In this context, commercial games have emerged as practical platforms for synthetic data generation \cite{richter2016playing}. 
    While not originally designed for this purpose, they are widely adopted due to their rich pre-built assets, which eliminate the substantial cost of asset creation required in general-purpose simulation platforms such as Unity \cite{unity} and Unreal \cite{unreal} Engine \cite{bala2024edify, yang2025genassets, blaga2025breaking}.
Among these, GTA V-based pipelines established a prominent line of work demonstrating large-scale automatic annotation for visual perception. 
    Prior studies leveraged GTA V \cite{gta5} and similar environments to construct datasets spanning multiple tasks like segmentation \cite{richter2016playing}, crowd counting \cite{wang2019learning}, anomaly detection \cite{lin2021learning}, showing that game-based pipelines can support scalable and diverse supervision. 
More recent efforts further extend this paradigm to specialized tasks such as scene flow estimation \cite{jin2022deformation} and HDR reconstruction \cite{barua2025gta}.
    % indicating continued progress in game-driven data generation.

% In parallel, CARLA \cite{dosovitskiy2017carla} has become a canonical programmable simulator for autonomous-driving research, supporting configurable sensors, environmental conditions, and standardized evaluation \cite{sun2022shift, xu2022safebench, jia2024bench2drive}. 

However, existing game-based data generation pipelines remain insufficient for RGB-T aerial object detection. 
    For instance, GTA V has mainly been used for visible-domain data generation, as it does not natively provide thermal observations. 
Arma3 is a large-scale tactical simulation game that provides open-world outdoor environments, controllable entities, and scriptable in-game cameras.
    Besides, Arma3 enables RGB-T paired data generation.
Nevertheless, Arma3 has not been fully explored as a structured data-generation platform for multi-view RGB-T aerial object detection.

Hence, we present \textbf{\textit{G-MAD}}, an open-source framework that transforms Arma3 \cite{arma3} into a structured data-generation pipeline for multi-view RGB-T aerial object detection. 
    G-MAD\footnote{It stands for ``a data \underline{G}enerator for \underline{M}ulti-view \underline{A}erial object \underline{D}etection using Arma3.''} supports scenario specification, synchronized RGB-T multi-view capture, and automatic object annotation using engine-level geometric information. 
Using G-MAD, we further construct \textbf{\textit{AMOD}}, a new multi-view RGB-T aerial detection benchmark for studying viewpoint robustness, RGB-T learning, and synthetic-to-real transfer.

\section{Architecture of G-MAD}

\begin{figure}
\centering
% \vspace{-0.3cm}
\includegraphics[width=8.3cm]{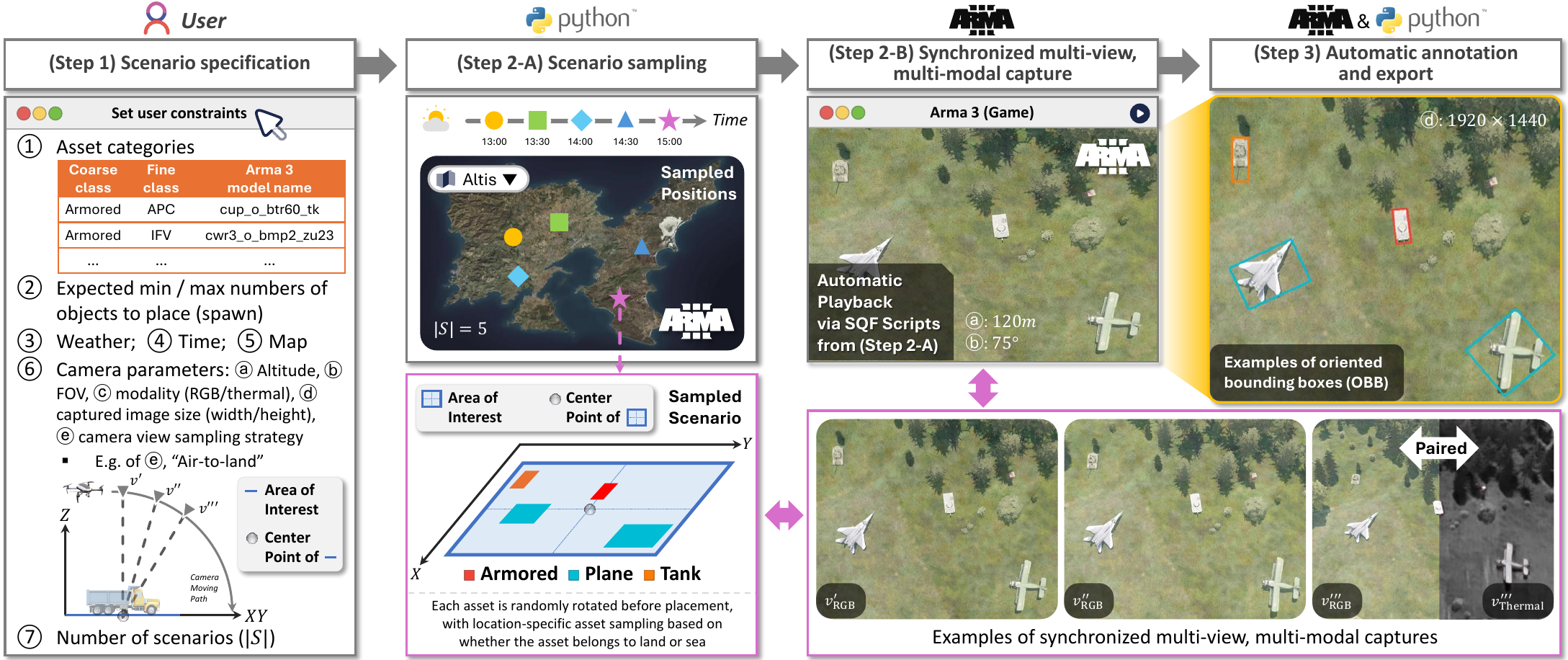}
  \vspace{-0.2cm}
  \caption{The overall architecture of G-MAD. }
  % \Description{Enjoying the baseball game from the third-base
  % seats. Ichiro Suzuki preparing to bat.}
  \label{gmad:fig:teaser}
\end{figure}

G-MAD is designed as an end-to-end data generation pipeline that transforms a compact set of user-specified parameters into structured multi-modal (RGB-T) data for aerial detection. 
    As illustrated in \textcolor{black}{Fig. \ref{gmad:fig:teaser}}, the proposed framework decomposes the data generation process into three stages: (1) scenario specification, (2) scenario sampling and synchronized multi-view, multi-modal capture, and (3) automatic annotation and export. 
The overall pipeline is implemented in SQF, the native scripting language of Arma3, allowing direct procedural control over scene composition, environmental configuration, in-game sensor control, and data export within the simulator.
In this work, we use a term \textit{\textbf{scenario}} to denote a parameterized specification of data generation conditions, including object arrangement and environmental factors such as time, weather, and background. A scenario defines a fixed underlying configuration shared across multiple observations. 
    Meanwhile, we use a term \textit{\textbf{scene}} to denote a concrete capture instance generated from a scenario, corresponding to a specific viewpoint and sensor configuration (e.g., camera pose and sensing modality like RGB). 
Under this formulation, a single scenario produces multiple scenes by varying viewpoints and sensing modalities while keeping the object arrangement and environment unchanged; see \textcolor{black}{Fig. \ref{gmad:fig:teaser}}.

\vspace{-0.3cm}
\subsection{Scenario specification} \label{scenario_specification}
% G-MAD formulates data generation as a parameter-driven process. 
    Rather than requiring users to manually construct individual scenarios, the proposed framework generates scenarios automatically from a compact set of high-level constraints provided by the user.
    % To enable \textbf{\textit{coarse-to-fine classification}}, the framework provides support for hierarchical class labeling. (E.g., Helicopter - Observing Helicopter)
These constraints include \ding{172} asset categories\footnote{We additionally provide a \textit{\textbf{pre-defined table that assigns each of the 513 unique Arma3 asset models}} to one of the 12 classes described in Sec. 4, enabling users to immediately test our code; kindly refer to ``classes/CLASSES.csv''. By referring to this table, users can create data for their desired assets. Moreover, users are \textit{\textbf{not limited to military assets}}; they can also import civilian equipment into Arma3. For further details, please refer to the ``README.md'' in our code and the official Arma3 website.}, \ding{173} the expected minimum and maximum numbers of objects to place, \ding{174} weather, \ding{175} time (day/night), \ding{176} map, \ding{177} camera sensor configuration (\ding{177}-a. Altitude, \ding{177}-b. FOV, \ding{177}-c. RGB-thermal modality\footnote{Basically, \textbf{\textit{white-hot thermal images}} are generated using the Arma3 TI engine. If specific spectrum (e.g., MWIR) is required, you may use image-to-image translation techniques (e.g., \cite{cheong2024thermal}), to convert the generated images into the target spectrum.}, \ding{177}-d. captured image size (width/height), \ding{177}-e. camera view sampling strategy per each scene), \ding{178} number of scenarios.
% \footnote{\textcolor{black}{Please see ``README.md'' of our code for details.}}
    % The scenario specification therefore defines only the admissible generation space rather than concrete object layouts or camera targets. 
Under these constraints, G-MAD automatically samples each scenario by randomly determining asset category, object arrangement, camera target point, and time, while keeping other factors like weather and sensor settings fixed.
    % For example, a user may specify that each generated scenario contains 5--10 non-overlapped assets within a predefined image resolution, under clear weather conditions, with a fixed camera altitude of 120\,m and a field of view of 75$^\circ$, using paired RGB-T captures from multiple viewing angles (20$^\circ$ and 40$^\circ$).

\begin{table}[t]
\centering
\caption{Comparison of Arma3-based open source synthetic data generation frameworks in terms of aerial detection.} \vspace{-0.2cm}
\label{gmad:tab:comparison}
\resizebox{8.3cm}{!}{
\begin{tabular}{lccc}
\toprule
 & \cite{armacode1} & \cite{armacode2} & G-MAD (Proposed) \\
\midrule

\multicolumn{4}{c}{\textit{\textbf{Scene configuration and sampling}}} \\
Multi-map support 
& $\times$ 
& $\times$ 
& \textbf{$\bigcirc$} \\

Max objects per image 
& $\leq$ 1 
& $\leq$ 1 
& \textbf{$\geq$ 1} \\

Region of interest selection
& Manual 
& Manual
& \textbf{Automatic (Constraint-aware)}
\\

Camera viewpoint sampling
& Predefined orbit
& Unconstrained random
& \textbf{User-defined sampling model} \\

% Weather variation
% & Fog, Sun, Rain
% & Fog, Sun, Rain
% & \textbf{Fog, Overcast, Sun, Rain, Snow} \\

Occlusion check between camera and object 
& $\times$ 
& $\bigcirc$ 
& \textbf{$\bigcirc$} \\

\midrule
\multicolumn{4}{c}{\textit{\textbf{Annotation and output}}} \\
Annotation method 
& Post-hoc image differencing
& Post-hoc image differencing
& \textbf{Engine-native ground truth} \\

Bounding box support 
& $\times$ 
& HBB 
& \textbf{HBB, OBB} \\

Resolution (Image size)
& Fixed 
& Fixed 
& \textbf{Configurable} \\

Resolution scaling (supersampling)
& $\times$
& $\times$
& \textbf{$\bigcirc$} \\

\midrule
\multicolumn{4}{c}{\textit{\textbf{Execution and automation}}} \\
MATLAB dependency 
& $\bigcirc$ 
& $\times$ (Only Python)
& \textbf{$\times$ (Only Python)} \\

Execution scheme
& GUI-dependent control
& GUI-dependent control
& \textbf{GUI-independent launch} \\

Temporary file cleanup &
Manual &
Manual &
\textbf{Automatic} \\

Workstation availability during generation
& Unavailable
& Unavailable
& \textbf{Available} \\

\bottomrule 
\end{tabular} 
} \vspace{-0.2cm}
\end{table}

\begin{table*}[!t]
\centering
\caption{Comparison of the AMOD dataset with existing RGB-T aerial object detection benchmarks.}
\label{amod:tab:comparison}
\vspace{-0.2cm}

\begin{minipage}[t]{0.57\textwidth}
\vspace{0pt}
\centering
\setlength{\tabcolsep}{4pt}
\renewcommand{\arraystretch}{1.08}
\resizebox{.88\linewidth}{!}{
\begin{tabular}{@{}lccccccccc@{}}
\toprule
Dataset & Scenario & \begin{tabular}[c]{@{}c@{}}Image\\Size\end{tabular} & \begin{tabular}[c]{@{}c@{}}Total\\Frames\end{tabular} & \begin{tabular}[c]{@{}c@{}}Bounding\\Box Type\end{tabular} & \begin{tabular}[c]{@{}c@{}}Total\\Instances\end{tabular} & \begin{tabular}[c]{@{}c@{}}Total\\Categories\end{tabular} & Location & ID & \begin{tabular}[c]{@{}c@{}}Simultaneous\\Viewpoints\end{tabular} \\
\midrule
MULTISPECTRAL (MM-17) \cite{takumi2017multispectral} & Driving (DV) & 640x480 & 7,512 & HBB & 5.8k & 5 & Asia & N & Single \\
FLIR\_ADAS (FLIR-18) \cite{flir} & DV & 480x640 & 14,000 & HBB & 81.7k & 4 & North America & N & Single \\
DroneVehicle (TCSVT-22) \cite{sun2022drone} & Drone (DR) & 640x512 & 56,878 & OBB & 88.3k & 5 & Asia & N & Single \\
FLIR\_ADAS\_v2 (FLIR-22) \cite{flir} & DV & 640x512 & 26,442 & HBB & 375k & 15 & North America & N & Single \\
MSRS (IF-22) \cite{tang2022piafusion} & DV & 480x640 & 2,888 & HBB & 5.4k & 8 & Asia & N & Single \\
M3FD (CVPR-22) \cite{liu2022target} & DR + DV & 1024x768 & 8,400 & HBB & 33.6k & 6 & Asia & N & Single \\
DVTOD (TIV-24) \cite{song2024misaligned} & DR & 1920x1080 & 4,358 & HBB & 12.2k & 3 & Asia & N & Single \\
SMOD (TMM-25) \cite{chen2025amfd} & DV & 640x512 & 17,352 & HBB & 31.4k & 4 & Asia & N & Single \\
ESCVehicle (TGRS-26) \cite{song2026escvehicle} & DR & 704x704 & 21,454 & OBB & 36.9k & 7 & Asia & N & Single \\
\rowcolor[HTML]{96FFFB}
\textbf{AMOD (Ours)} & \textbf{DR} & \textcolor{red}{\textbf{1920x1440}} & \textcolor{red}{\textbf{73,920}} & \textbf{OBB} & \textcolor{red}{\textbf{383.2k}} & \textbf{12} & \textcolor{red}{\textbf{Multiple}} & \textcolor{red}{\textbf{Y}} & \textcolor{red}{\textbf{Multiple}} \\
\bottomrule
\end{tabular}
}
\end{minipage}
\hspace{0.00\textwidth}
\begin{minipage}[t]{0.33\textwidth}
\vspace{-0.15cm}
\centering
\includegraphics[
    width=0.7\linewidth,
]{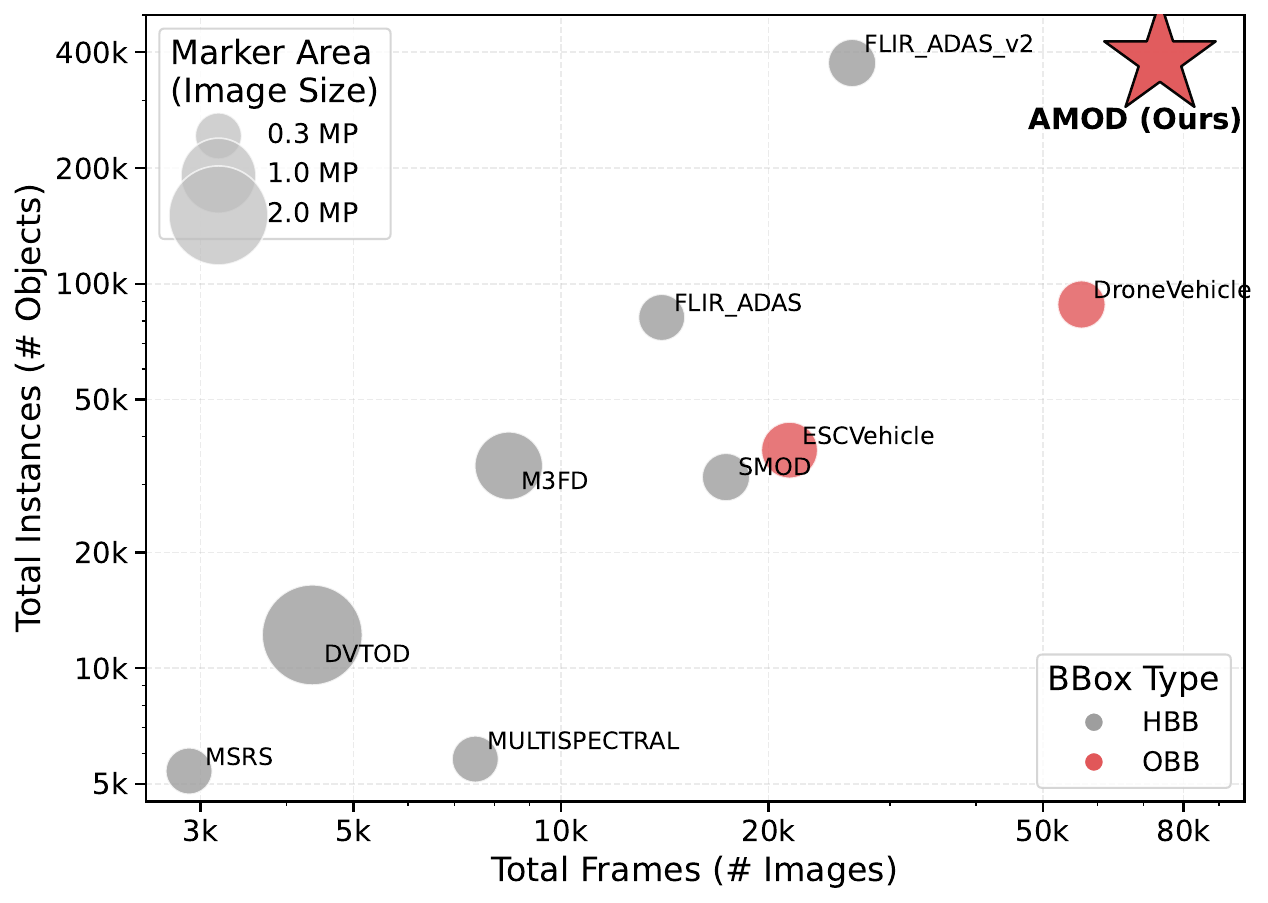}
\end{minipage}

\vspace{-0.2cm}
\end{table*}

\begin{figure}[!t]
\vspace{-0.0cm}
  \includegraphics[width=7cm]{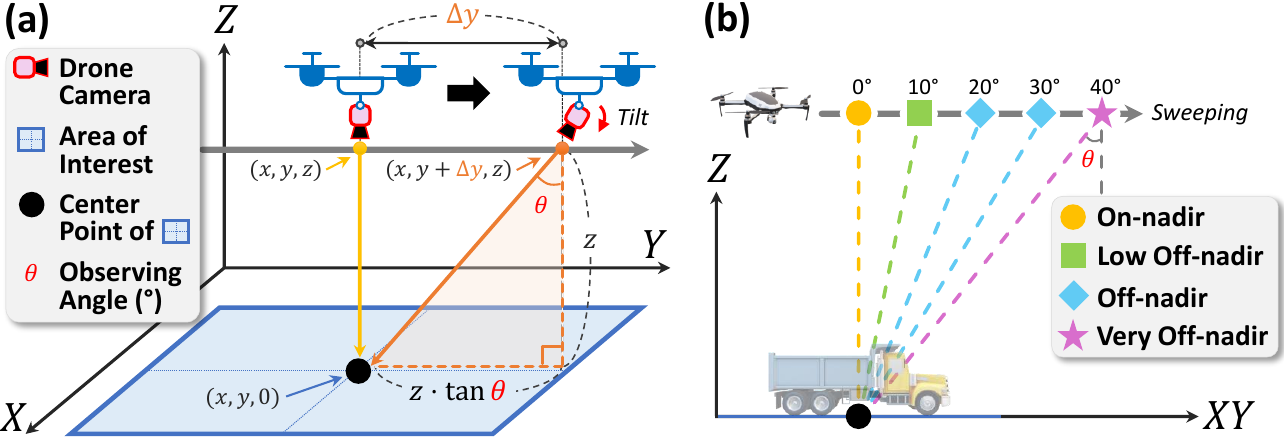}
  \caption{Illustration of one of the default camera viewpoint sampling methods in G-MAD: ``air-to-air''.
(a) Geometric parameterization of camera motion around the area of interest, defined by the observing angle and lateral displacement.
(b) Viewpoint sweep from nadir to off-nadir.}
% (c) Example captured images corresponding to the sampled viewpoints.}
  % \Description{Enjoying the baseball game from the third-base
  % seats. Ichiro Suzuki preparing to bat.}
  \label{gmad:fig:2} \vspace{-0.2cm}
\end{figure}

\vspace{-0.2cm}
\subsection{Scenario sampling and multi-view, multi-modal capture}

Given the user-specified constraints described in Sec. \ref{scenario_specification}, G-MAD formulates data generation as a two-stage sampling process over scenarios and viewpoints.
    Let $\mathcal{S}$ denote the space of valid scenarios defined by user constraints, and $\mathcal{V}$ denote the space of possible viewpoints and sensing configurations. The data generation process can be expressed as $s\sim P(\mathcal{S} \mid \text{constraints}),\; v \sim P(\mathcal{V} \mid s)$,
% \end{equation}
where each sampled pair $(s, v)$ produces a single annotated scene.

This formulation decouples scenario-level variables (object arrangement, environment) from scene-level variables (camera viewpoint and sensing modality), enabling structured data generation.
    A single scenario can produce multiple consistent observations by varying viewpoints and sensing modalities while preserving the underlying layout.
Specifically, for each sampled scenario, G-MAD generates multiple scenes by sampling camera poses around a target region and assigning sensing modalities such as RGB and thermal. 
    This yields a structured dataset of size $|\mathcal{S}| \times |\mathcal{V}|$, where scenes derived from the same scenario share identical object configurations but differ in viewpoint and modality.

    Camera viewpoints are defined through a user-configurable sampling model, allowing \textit{controlled multi-view data generation}.
The current G-MAD framework offers two default motion modes, ``air-to-air'' and ``air-to-land'' for users who prefer not to design a custom sampling model.
    In ``air-to-air'', the camera moves laterally while maintaining a relatively stable altitude, producing sweeping aerial observations; see Fig. \ref{gmad:fig:2}. 
In ``air-to-land'', both altitude and lateral position vary following a parabolic trajectory, resulting in oblique views directed toward the ground\textcolor{black}{; see Fig. \ref{gmad:fig:teaser}}.

Importantly, the viewpoint sampler is designed to be modular and extensible\footnote{Users can implement their own camera viewpoint sampling model in ``src/my\_scene\_generator/util.py". Specifically, please carefully update ``get\_all\_cam\_and\_target\_points()'' and ``get\_all\_cam\_and\_target\_points\_diff()''. \textcolor{black}{For more details, please see ``README.md'' of our code.}}. 
    Rather than relying on a fixed trajectory, G-MAD defines camera poses through a parameterized sampling routine, allowing users to incorporate custom motion patterns with minimal modification, e.g., circular orbits or task-specific reconnaissance paths.
Such flexibility enables users to tailor viewpoint distributions to specific downstream tasks.

This structured generation process provides two key benefits for learning. 
        First, it enables consistent supervision across viewpoints, allowing models to learn viewpoint-invariant representations from multiple observations of the same underlying scene. 
    Second, it naturally supports aligned multi-modal data generation, where RGB and thermal images share identical annotations, facilitating supervised cross-modal and fusion-based learning.

% \begin{figure}[!t]
%   \includegraphics[width=8.3cm]{Figs/AMOD/Fig2.pdf} \vspace{-0.2cm}
%   \caption{Annotated examples from the AMOD dataset featuring oriented bounding boxes (OBB). The dataset includes paired RGB-T imagery (see region `0456/10$^\circ$' for instance).}
% % (c) Example captured images corresponding to the sampled viewpoints.}
%   % \Description{Enjoying the baseball game from the third-base
%   % seats. Ichiro Suzuki preparing to bat.}
%   \label{gmad:fig:3} \vspace{-0.2cm}
% \end{figure}

% \begin{figure}[!t]
%   \includegraphics[width=6.5cm]{Figs/GMAD/Fig4.pdf} \vspace{-0.2cm}
%   \caption{Qualitative analysis of an AMOD-trained detector on unseen real-world RGB imagery containing military equipment on real-world RGB satellite photographs released through public news coverage of the Russia–Ukraine war.}
% % (c) Example captured images corresponding to the sampled viewpoints.}
%   % \Description{Enjoying the baseball game from the third-base
%   % seats. Ichiro Suzuki preparing to bat.}
%   \label{gmad:fig:4} \vspace{-0.2cm}
% \end{figure}

\vspace{-0.2cm}
\subsection{Automatic annotation for object detection}

G-MAD implements a direct annotation pipeline by querying object-level geometric information from the Arma3 engine and projecting it into the image plane.
    Unlike prior approaches that rely on foreground-background image differencing, our framework retrieves precise 3D bounding box information from the Arma3's engine and transforms it into 2D annotations aligned with each captured scene, ensuring high-quality labels.
        % without manual intervention.
        % ; \textcolor{purple}{see Supp. Material for details}.

Formally, let an object in a scenario be associated with a set of 3D bounding box vertices. 
    These vertices are projected onto the image plane using the camera parameters corresponding to each sampled viewpoint. 
The resulting 2D coordinates are used to generate both horizontal and oriented bounding boxes (HBB/OBB), providing richer supervision for detection tasks.

Furthermore, G-MAD incorporates automatic filtering of invalid annotations, such as objects fully occluded by other scene elements, as determined by the absence of a direct line-of-sight between the camera and the object surface.
% ; \textcolor{purple}{see Supp. Material for details}.
    This reduces label noise of ghost bounding boxes compared to pipelines such as \cite{armacode1} that do not explicitly handle occlusion.

\vspace{-0.1cm}
\subsection{Comparison with existing Arma3-based open source data generation frameworks}

To clarify the contribution of G-MAD, we compare it with two representative open-source Arma3-based data generation frameworks in Tab.~\ref{gmad:tab:comparison}. 
    Prior approaches~\cite{armacode1,armacode2} are limited in both data diversity and systematic generation: they assume at most one object per image, rely on manual or inflexible camera setup, and provide little support for research-oriented, controlled dataset construction. 
Another key limitation is execution control. Existing methods depend on GUI-level automation via click-based actions to launch and operate Arma3, making the pipeline brittle to UI state changes and user interference. 
    In addition, their annotations are obtained through image differencing\footnote{In \cite{armacode1, armacode2}, the object localization is inferred indirectly from image differencing between two distinct renders, one with the object present and the other with the object absent.}, which is cumbersome and error-prone for multi objects.
        Furthermore, this inherently limits its extensibility to data generation frameworks for video-level tasks, such as tracking.

In contrast, G-MAD enables constraint-aware multi-object generation, GUI-independent execution, and Arma3 engine-native annotation of both HBBs and OBBs. 
    Unlike \cite{armacode1, armacode2}, our work further supports multi-map backgrounds, configurable resolution with supersampling, and automatic cache cleanup, establishing Arma3 as a structured and scalable data generation framework for object detection in overhead imagery.

\vspace{-0.2cm}
\section{Application of G-MAD} \label{app_gmad}
To demonstrate the research utility of G-MAD, we highlight three representative application settings enabled by our framework: 1) multi-view learning, 2) cross-modal RGB-T learning and fusion, and 3) pretraining for real-world object detection.
    Because G-MAD generates multiple synchronized observations from a shared underlying scenario while allowing controlled variation of viewpoint, modality, and environmental factors, it provides a practical foundation for studying data-centric learning problems.

\vspace{-0.2cm}
\section{Benchmarking test of G-MAD}

\begin{figure}
\centering
\vspace{-0.3cm}
\includegraphics[width=6cm]{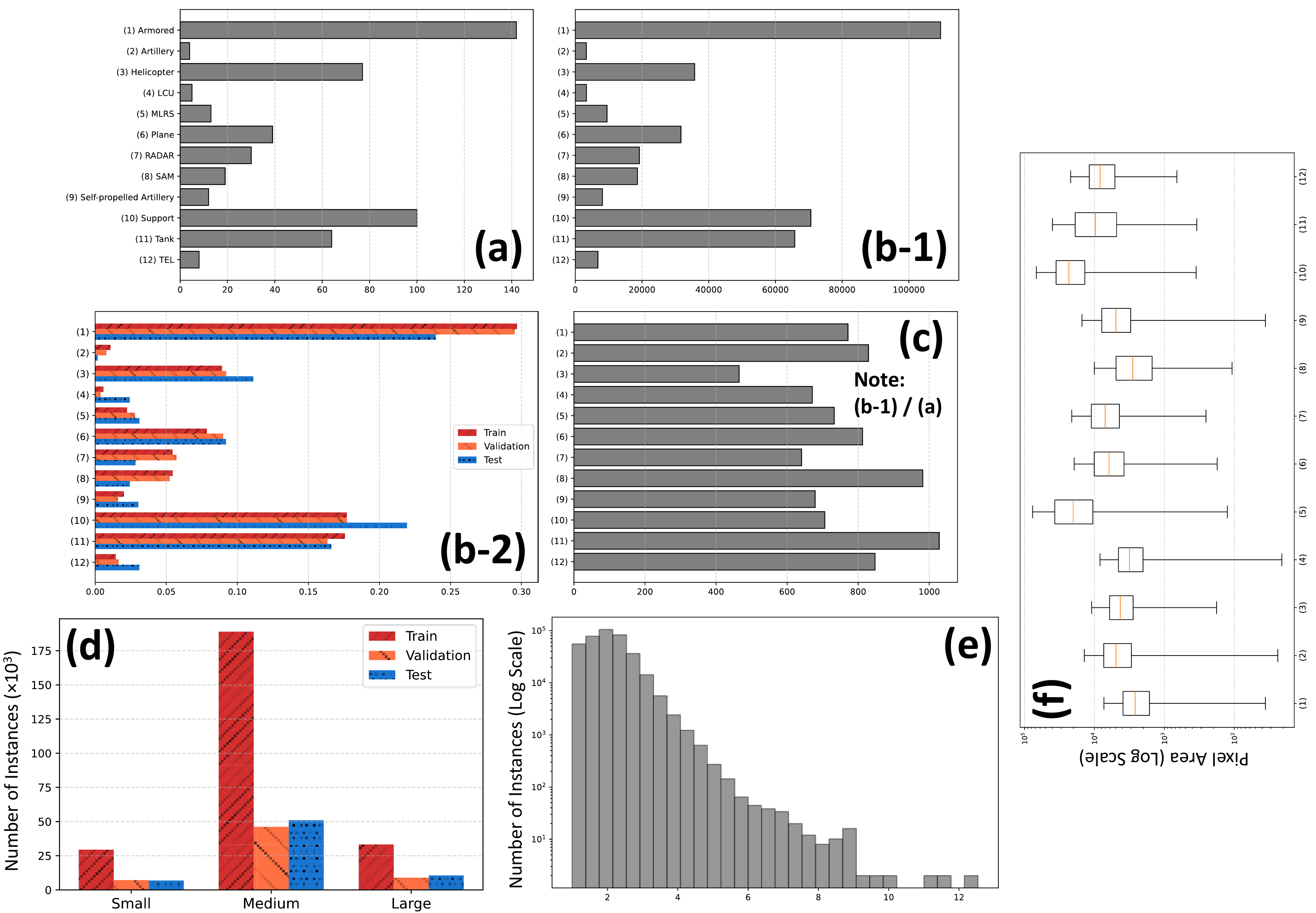}
  \vspace{-0.2cm}
  \caption{Statistics of the AMOD dataset. (a) Object models per category used in Arma3. (b-1) Instance count per category. (b-2) Instance distribution per category in train/dev/test splits. (c) Normalized instance count by Arma3 models across categories. (d) Size distribution (Small/Medium/Large) in train/dev/test splits. (e) Instance size of OBBs per category. (f) Aspect ratio distribution of OBBs per category. 
  % Note that RGB and thermal modalities are perfectly aligned, with identical image counts, instance counts, and bounding box annotations in our benchmark.
  } \vspace{-0.2cm}
  % \Description{Enjoying the baseball game from the third-base
  % seats. Ichiro Suzuki preparing to bat.}
  \label{amod:fig:3}
\end{figure}

\begin{figure}[!t]
  \includegraphics[width=8.3cm]{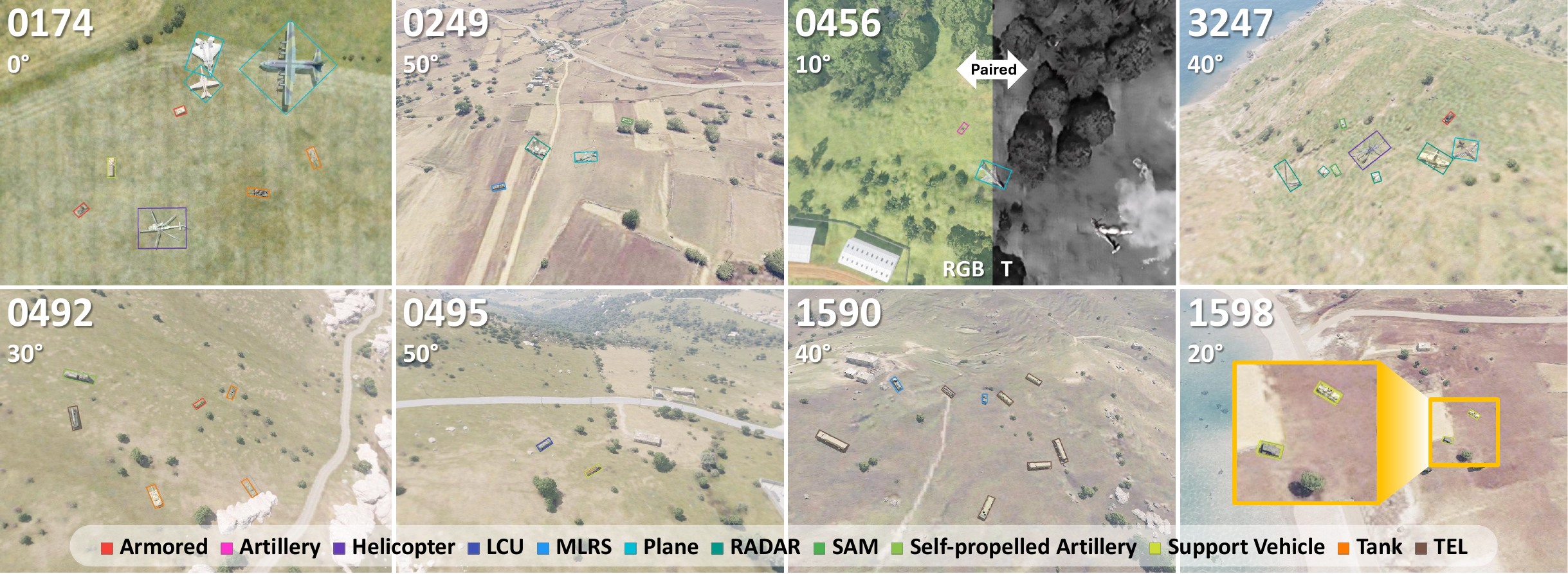} \vspace{-0.2cm}
  \caption{Annotated examples from the AMOD dataset featuring oriented bounding boxes (OBB). The dataset includes paired RGB-T imagery (see region `0456/10$^\circ$' for instance).}
% (c) Example captured images corresponding to the sampled viewpoints.}
  % \Description{Enjoying the baseball game from the third-base
  % seats. Ichiro Suzuki preparing to bat.}
  \label{gmad:fig:3} \vspace{-0.2cm}
\end{figure}

\begin{figure}[!t]
  \includegraphics[width=6.5cm]{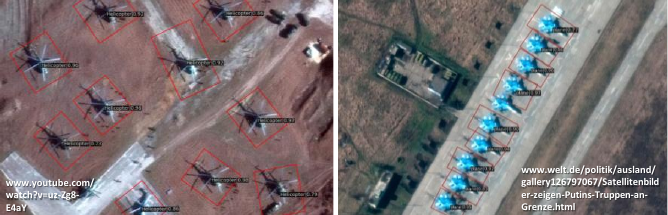} \vspace{-0.2cm}
  \caption{Qualitative analysis of an AMOD-trained detector on unseen real-world RGB imagery containing military equipment on real-world RGB satellite photographs released through public news coverage of the Russia–Ukraine war.}
% (c) Example captured images corresponding to the sampled viewpoints.}
  % \Description{Enjoying the baseball game from the third-base
  % seats. Ichiro Suzuki preparing to bat.}
  \label{gmad:fig:4} \vspace{-0.2cm}
\end{figure}

To validate the effectiveness of G-MAD, we conduct a set of benchmarking experiments focusing on two aspects, 1) and 3), aligned with Sec. \ref{app_gmad}. We construct a dataset using G-MAD, referred to as Aerial Military Object Detection (AMOD), which consists of synchronized RGB-T multi-view aerial observations with OBB annotations of following categories: \textit{Armored}, \textit{Artillery}, \textit{Helicopter}, \textit{LCU}, \textit{MLRS}, \textit{Plane}, \textit{RADAR}, \textit{SAM}, \textit{Self-propelled Artillery}, \textit{Support Vehicle}, \textit{Tank}, and \textit{TEL}; see Figs. \ref{amod:fig:3}, \ref{gmad:fig:3}, Tab. \ref{amod:tab:comparison}, and Supp. Material for details. 
    As a baseline detector, we adopt Oriented R-CNN \cite{xie2024oriented} with a Swin-S \cite{liu2021swin} backbone, following prior work \cite{kim2025nbbox, lei2025mgnet, wang2026adaptive, zheng2026see} on aerial detection, using MMRotate \cite{zhou2022mmrotate}.

\noindent\textbf{Multi-view learning.} 
We first evaluate the effect of multi-view supervision using synchronized observations generated from the same scenarios. As shown in Tab.~\ref{gmad:tab:2}, while single-view models degrade under unseen viewing angles, \textit{multi-view training} significantly improves generalization, achieving a mean AP$_{50:95}$ of 50.96 compared to 44.57 for the best single-view model (+6.39). 
    % Moreover, for fair comparison with single-views in terms of data quantity, when the all-view training set is reduced to one-sixth of its original size, the resulting model still attains 47.61 mean AP, outperforming the best single-view model by +5.86.
    % This suggests that viewpoint diversity can be a major contributor to object detection.

\begin{table}[]
\centering
\caption{Cross-angular evaluation results (AP$_{50:95}$) on AMOD.} \vspace{-0.4cm}
\label{gmad:tab:2}
\resizebox{4.4cm}{!}{%
\begin{tabular}{@{}cccccccc@{}}
\toprule
 & \multicolumn{7}{c}{Test} \\ \cmidrule(l){2-8} 
Train & \multicolumn{1}{c}{0$^\circ$} & \multicolumn{1}{c}{10$^\circ$} & \multicolumn{1}{c}{20$^\circ$} & \multicolumn{1}{c}{30$^\circ$} & \multicolumn{1}{c}{40$^\circ$} & \multicolumn{1}{c}{50$^\circ$} & \multicolumn{1}{c}{Mean} \\ \midrule
0$^\circ$ & \underline{51.16} & 49.62 & 47.44 & 43.07 & 36.88 & 26.60 & 42.46 \\
10$^\circ$ & 50.92 & \underline{50.01} & 48.04 & 44.79 & 38.76 & 28.39 & 43.49 \\
20$^\circ$ & 50.51 & 49.88 & \underline{48.57} & 45.64 & 40.13 & 32.71 & \underline{44.57} \\
30$^\circ$ & 48.67 & 48.27 & 47.42 & \underline{45.69} & 41.46 & 33.66 & 44.20 \\
40$^\circ$ & 46.56 & 46.67 & 46.41 & 44.80 & \underline{42.24} & \underline{37.07} & 43.96 \\
50$^\circ$ & 38.23 & 38.20 & 39.04 & 39.29 & 37.95 & 36.49 & 38.20 \\  \midrule
All & \textbf{56.18} & \textbf{54.75} & \textbf{53.93} & \textbf{51.03} & \textbf{47.67} & \textbf{42.20} & \textbf{50.96} \\
 & (\textcolor{red}{+5.02}) & (\textcolor{red}{+4.74}) & (\textcolor{red}{+5.36}) & (\textcolor{red}{+5.34}) & (\textcolor{red}{+5.43}) & (\textcolor{red}{+5.13}) & (\textcolor{red}{+6.39}) \\  
 % \midrule 
% All (Reduced to $\frac{1}{6}$) & \textbf{56.92} & \textbf{52.51} & \textbf{53.22} & \textbf{50.84} & \textbf{39.74} & \textbf{32.44} & \textbf{47.61} \\ 
 % & (\textcolor{red}{+3.21}) & (\textcolor{red}{+2.01}) & (\textcolor{red}{+5.41}) & (\textcolor{red}{+6.43}) & (\textcolor{red}{+2.61}) & (\textcolor{red}{+3.91}) & (\textcolor{red}{+5.86}) \\ 
\bottomrule
\end{tabular} 
}  \vspace{-0.0cm}
\end{table} 

\begin{table}[]
\vspace{-0.0cm}
\centering
\caption{Investigating the effectiveness of pretraining with AMOD on two real-world benchmarks, DIOR-R (Visible) and HIT-UAV (Thermal) (AP$_{50}$).}
% \caption{AMOD vs. ImageNet pretraining on two real-world benchmarks, DIOR-R (Visible) and HIT-UAV (Thermal).}
\vspace{-0.4cm}
\label{gmad:tab:3}
\resizebox{7.6cm}{!}{%
\begin{tabular}{@{}l|ccccccccc@{}}
\toprule
\multirow{2}{*}{Pretrained from} & \multicolumn{5}{c|}{Finetuned for DIOR-R}                           & \multicolumn{4}{c}{Finetuned for HIT-UAV}            \\ \cmidrule(l){2-10} 
                            & Airplane & Ship  & Vehicle & Windmill & \multicolumn{1}{c|}{Mean}  & Person & Car   & Others & \multicolumn{1}{c}{Mean}  \\ \midrule
ImageNet                    & 71.88    & 81.02 & 42.59   & 57.07    & \multicolumn{1}{c|}{63.14} & 85.30  & 81.29 & 57.12  & \multicolumn{1}{c}{74.57} \\
AMOD                        & \textbf{79.71}    & \textbf{89.04} & \textbf{53.21}   & \textbf{57.36}    & \multicolumn{1}{c|}{\textbf{69.83}  (\textcolor{red}{+6.69})} & \textbf{87.68}  & \textbf{82.15} & \textbf{61.40}  & \textbf{77.08} (\textcolor{red}{+2.51})                      \\ \bottomrule 
\end{tabular}
} \vspace{-0.3cm}
\end{table}

\noindent\textbf{Pretraining for real-world detection.}  
We further evaluate a synthetic-to-real transfer setting, where models are pretrained on RGB and thermal variants of AMOD and fine-tuned on real-world RGB and thermal benchmarks, respectively.
    As shown in Tab.~\ref{gmad:tab:3}, AMOD pretraining consistently outperforms ImageNet initialization, improving mean AP$_{50}$ from 63.14 to 69.83 on DIOR-R (+6.69) and from 74.57 to 77.08 on HIT-UAV (+2.51). 
This suggests that AMOD provides transferable pretraining signals for both visible and thermal real-world aerial detection\footnote{\textcolor{black}{For pretraining, we perform a one-time syn-to-real style transfer of our AMOD toward the target RGB/thermal domain, respectively; the detailed pipeline is in Supp. Material.}}.

\noindent\textbf{Unseen real-world inference of military targets.} Since authoritative real-world datasets for military target detection are difficult to obtain, we further perform an unseen real-world inference test from public news coverage, strictly without any fine-tuning.
    As shown in Fig.~\ref{gmad:fig:4}, the AMOD-trained detector\footnote{During AMOD training, only the RGB split is leveraged, incorporating DOTA \cite{xia2018dota}-style transfer; kindly refer to Supp. Material for details.} still identifies plausible military targets on previously unseen real images.

\noindent\textbf{Limitations of AMOD.}
Although AMOD provides a large-scale, synchronized multi-view RGB-T benchmark for aerial object detection, it is primarily specialized for military targets. 
    This specialization stems from the use of Arma3, whose built-in assets and terrains are particularly well suited for military scenarios. 
        However, since Arma3 supports user-defined asset integration, G-MAD can also be used to construct datasets for non-military general equipment.

% Another limitation is the scope of the experimental validation reported in the main paper. For compactness, we present results using a representative oriented object detector, Oriented R-CNN, as the baseline model. In our additional experiments with various detection models, we observed similar trends to those reported in the paper, including the benefit of multi-view supervision and the effectiveness of AMOD pretraining. We will provide these extended model-wise results in the supplementary material.

\vspace{-0.2cm}
\section{Conclusions}
% In this work, we presented G-MAD, a scalable and configurable framework for generating synthetic RGB-T military asset detection data with synchronized multi-view capture and automatic annotation. 
%     Preliminary experiments suggest that our work supports both controlled benchmarking and transfer to real-world settings.
% As future work, we plan to extend G-MAD into a more advanced data generation platform for UAV+UGV detection scenarios.  
%     We also aim to broaden the framework beyond detection toward more challenging tasks, including segmentation and tracking.
In this work, we presented G-MAD, a scalable and configurable framework for generating synthetic RGB-T aerial detection data with synchronized multi-view capture and automatic annotation.
    Using G-MAD, we also constructed and released AMOD, an RGB-T aerial multi-view object detection dataset, and preliminary experiments with AMOD suggest that the proposed framework supports both controlled benchmarking and transfer to real-world settings.

\noindent\textbf{Licensing and ethical considerations.}
% Regarding licensing, G-MAD is designed for non-commercial academic research. It does not redistribute the Arma3 executable, extracted game files, or original asset packages; instead, it uses a legitimately obtained copy of Arma3 to render synthetic images and generate annotations. Users are responsible for obtaining a valid Arma3 license and complying with the applicable terms when using or extending the framework.
G-MAD relies on Arma3 as an off-the-shelf rendering and scenario-execution environment. 
    Accordingly, the framework is designed to avoid redistributing Bohemia Interactive intellectual property including Arma3 executable game files, original asset packages, or modified copies of the game. 
G-MAD only provides external scripts that users may run with their own legitimately obtained local installation of Arma3. 
    The framework is released solely for non-commercial academic or prototype research purposes, and users are responsible for ensuring that their use complies with the Arma3 End User License Agreement.

\begin{acks}
For the development of commercial software, LIG Defense\&Aerospace (LIG D\&A) only uses duly licensed subscriptions to VBS4, a professional military simulation platform developed by Bohemia Interactive Simulations. Arma3, developed by Bohemia Interactive, was used in this study solely for non-commercial academic purposes to facilitate the publication and dissemination of the paper and related research outcomes to the broader research community. It was not used in the development of the company’s commercial software. Moreover, the views, analyses, and conclusions presented in this paper are solely those of the authors and do not represent the official position, policies, or endorsement of LIG D\&A.
    This work was partly supported by the National Research Foundation of Korea (NRF) Grant by the Korean Government through Ministry of Science and ICT (MSIT) under Grant No. RS-2026-25475324, and GIST SCENT (A100). 
    % This work also benefited from high-performance GPU resources provided by HPC-AI Open Infrastructure via GIST SCENT (A100).
The authors would like to thank SooYeon Kim (KRIT), Jongmin Joo (GIST), Sung Heon Kim (GIST), Sihyun Kim (YU), Junggyun Oh (HDC), and Yeongmin Ko (KNU) for their assistance.
    Additional experimental results and a more detailed description of our AMOD dataset are provided in Supp. Material.
    
\end{acks}

\title{Supplementary Material for\\ ``G-MAD: A Game-Based Data Generation Framework for Multi-View RGB-T Aerial Object Detection''}

\setcounter{table}{0}
\setcounter{figure}{0}
\setcounter{section}{0}

\renewcommand{\thetable}{S-\arabic{table}}
\renewcommand{\thefigure}{S-\arabic{figure}}

\renewcommand{\tablename}{Table}
\renewcommand{\figurename}{Figure}

\renewcommand{\thesection}{§\Alph{section}}
\renewcommand{\thesubsection}{\thesection.\arabic{subsection}}

\begin{figure*}[!t]
    \centering
    \includegraphics[angle=0, width=17.5cm]{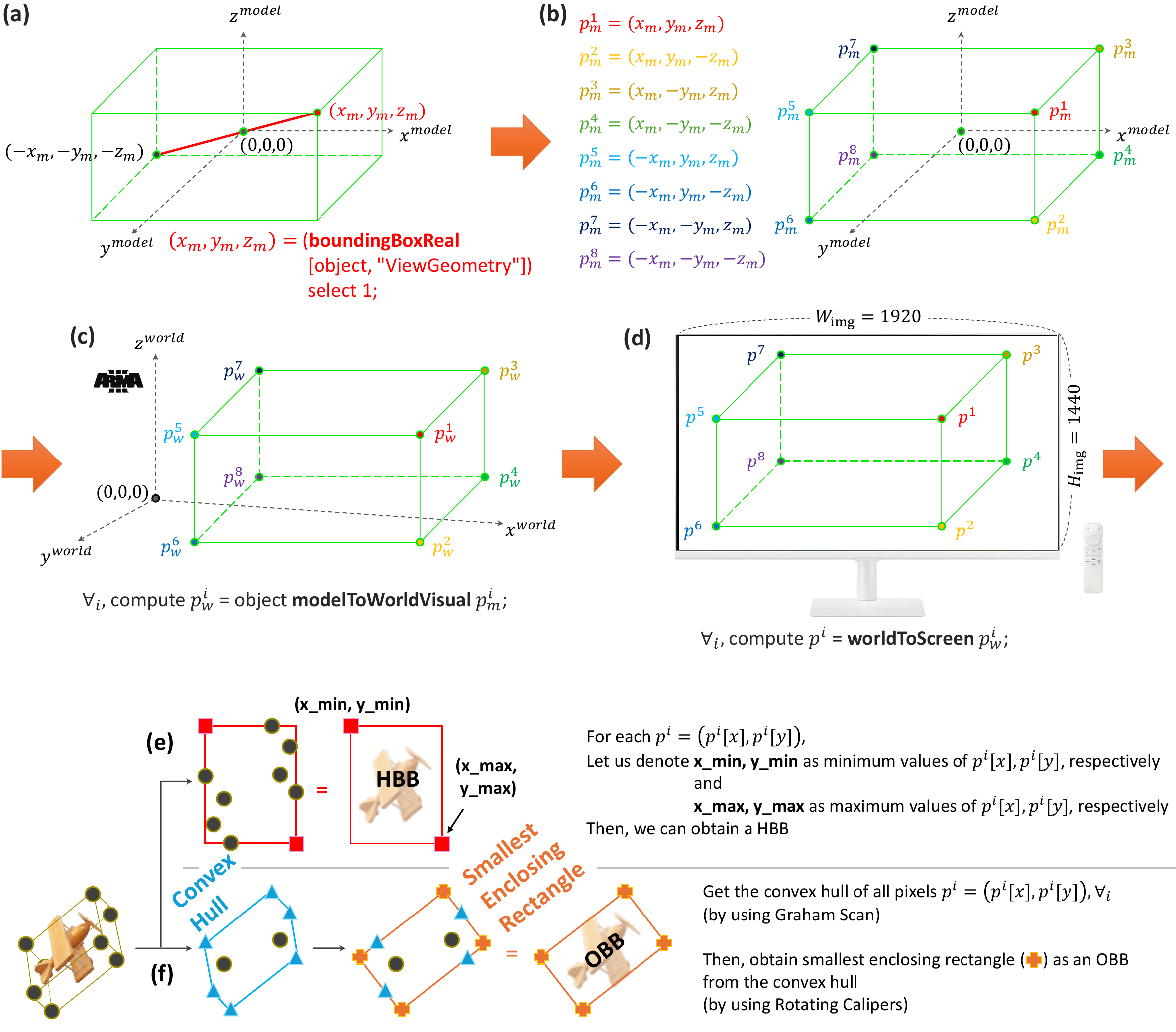} \vspace{-0cm}
    \caption{Overview of the pipeline for computing both a 2D horizontal bounding box (HBB) and an oriented bounding box (OBB) for a certain 3D object in Arma3. (a) The `\textbf{boundingBoxReal}' function is called in Arma3, which returns the lower and upper corners of the object's bounding box (BBox) in the object's own (local) model coordinate system ($m$), assuming the object itself is located at the origin; selecting index 1 yields the upper-corner coordinate $(x_m, y_m, z_m)$. (b) Using $(x_m, y_m, z_m)$, all eight $m$-space corner points $p^{i}_{m}$ ($i=1, \cdot\cdot\cdot, 8$) are generated by enumerating all sign combinations of $\pm x_m, \pm y_m,$ and $\pm z_m$. (c) Each model-space corner is transformed into the (global) Arma3 world coordinate frame ($w$) by invoking the `\textbf{modelToWorldVisual}' function of Arma3, producing $p^{i}_{w}$. (d) The $w$-space points are then projected into the image plane frame via the Arma3's `\textbf{worldToScreen}' function, yielding $p^{i} \in \mathbf{R}^{2}$. (e) From these projected points, the HBB is obtained in screen coordinates by taking the minimum and maximum from $x$ and $y$ values in $\{p^{i}=(p^{i}[x], p^{i}[y]) \}_{i=1}^{8}$. (f) The convex hull of all projected points is computed using Graham Scan, and the smallest enclosing rectangle of this hull is found using the Rotating Calipers algorithm, resulting in the final 2D OBB in screen coordinates.}
    \label{supfig1}
    \vspace{-0.0cm}
\end{figure*}

\section{{Bounding box extraction from Arma3}}
This section describes the procedure used to automatically extract 2D bounding boxes (BBoxes), both horizontal bounding boxes (HBBs) and oriented bounding boxes (OBBs), from 3D objects in Arma3. For convenience, we assume the BBox extraction for a single object.

\subsection{Obtaining the model-space BBox}

For each target object, we first invoke the function \textbf{\textit{boundingBoxReal}}, which returns the lower and upper corners of the object’s 3D BBox in the object’s (local) model coordinate system. Note that Arma3 assumes that each object is located at the origin of its own model space ($m$). Selecting index 1 yields the upper-corner coordinate $(x_{m}, y_{m}, z_{m})$; see Fig. \ref{supfig1}(a).
    Given this upper-corner value, we enumerate all sign combinations of $x_{m}$, $y_{m}$, and $z_{m}$ to generate the eight canonical model-space corner points $\{ p_{m}^{i} \}_{i=1}^{8}$; see Fig. \ref{supfig1}(b). 

\subsection{Transforming to world coordinates}

Each model ($m$)-space corner point is then mapped into the (global) Arma3 world coordinate frame ($w$) using the transformation function \textbf{\textit{modelToWorldVisual}}, producing world-space coordinates $p_{w}^{i}$ for $i\in \{1,\cdot\cdot\cdot,8\}$; see Fig. \ref{supfig1}(c).

\subsection{Projecting into screen coordinates}

To obtain 2D pixel-space coordinates, each world ($w$)-space point is projected onto the rendered image plane via the Arma3 camera projection function \textbf{\textit{worldToScreen}}; see Fig. \ref{supfig1}(d). 
    Though Fig. \ref{supfig1}(d) does not explicitly depict it, an additional post-processing step is required to obtain the actual pixel coordinates from \textbf{\textit{worldToScreen}} in Arma3.
The function \textbf{\textit{worldToScreen}} returns normalized screen coordinates expressed within the Arma3’s \textit{safe-zone} layout rather than the final pixel grid of the rendered image. 
    Consequently, one must apply safe-zone compensation before obtaining valid pixel-space positions as follows.

Let $(q^{i}[x], q^{i}[y])$ denote the 2D coordinates returned by \textbf{\textit{worldToScreen}}. The final pixel coordinates $(p^{i}[x], p^{i}[y])$ are computed by removing the safe-zone offset (SafeZoneX, SafeZoneY) and scaling with respect to the safe-zone width (SafeZoneW) and height (SafeZoneH):

\begin{equation}\label{eqsp1} \small
\setlength{\belowdisplayskip}{2pt}
\begin{split}
    p^{i}[x] = \frac{q^{i}[x]-\text{SafeZoneX}}{\text{SafeZoneW}}\times W_{\text{img}};\\
    p^{i}[y] = \frac{q^{i}[y]-\text{SafeZoneY}}{\text{SafeZoneH}}\times H_{\text{img}},\\
     % f(x; \sigma) = x \odot (1-{m}_{b}) + (x*{k}_{\sigma}) \odot {m}_{b}, \\
\end{split} 
\vspace{-0.5cm}
\end{equation} 

where SafeZoneX, SafeZoneY, SafeZoneW, and SafeZoneH are constants in Arma3; see {\small\url{https://community.bistudio.com/wiki/safeZone}} for details.
    In this work, we set $W_{\text{img}}$ and $H_{\text{img}}$ to 1920 and 1440, respectively, thus generating images with a resolution of 1920$\times$1440.

\subsection{Computing HBB and OBB in screen space}

Given the projected points $\{ p^{i} \}_{i=1}^{8}$, we compute both forms of BBox as follows:

\begin{itemize}
    \item \textbf{Horizontal Bounding Box (HBB).}  
    The HBB is obtained by taking the minimum and maximum of the $x$- and $y$-coordinates across all projected points; see Fig. \ref{supfig1}(e).

    \item \textbf{Oriented Bounding Box (OBB).}  
    The OBB is obtained by first computing the convex hull of the projected points using the Graham Scan algorithm. The smallest enclosing rectangle of this convex hull is then found using the Rotating Calipers method, yielding the final 2D OBB in screen coordinates; see Fig. \ref{supfig1}(f).
\end{itemize}

\subsection{Summary}

This pipeline enables accurate extraction of both HBB and OBB annotations directly from Arma3’s internal geometry and rendering functions. Because the pipeline relies solely on engine-grounded transformations and camera projections, the resulting annotations are geometrically consistent and suitableas ground-truth labels for training and evaluating aerial object detection models.

\section{AMOD: a novel synthetic benchmark for RGB-T aerial military object detection}

\begin{figure}[!ht]
\centering
\vspace{-0.0cm}
\includegraphics[width=8.1cm]{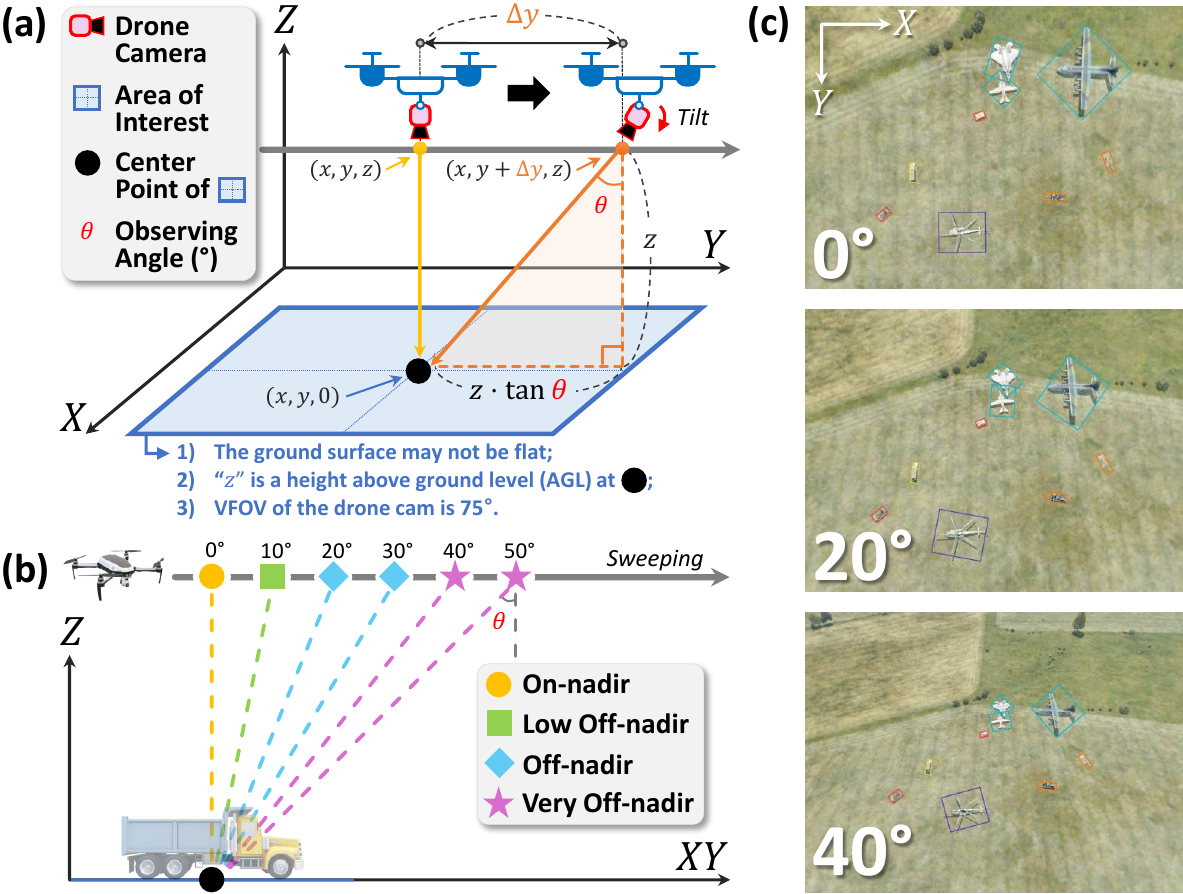}
  \vspace{-0.2cm}
  \caption{Illustration of how to construct our AMOD benchmark. (a) Geometric definition of the observing angle \textcolor{red}{$\theta$}, where the camera shifts laterally by \textcolor{orange}{$\Delta y$} in $Y$-axis, producing an off-nadir view. (b) Drone camera sweeping setup simultaneously capturing the same area from nadir to very off-nadir viewpoints. (c) Examples from the RGB version of AMOD under different observation angles, showing how target appearance and perspective distortion vary with \textcolor{red}{$\theta$}.}
  % \Description{Enjoying the baseball game from the third-base
  % seats. Ichiro Suzuki preparing to bat.}
  \label{amod:fig:teaser} \vspace{-0.2cm}
\end{figure}

We introduce \textbf{\textit{AMOD}}\footnote{It stands for ``\underline{A}erial \underline{M}ilitary \underline{O}bject \underline{D}etection in multi-view RGB-T scenarios.''}, a new benchmark to provide synchronized multi-view aerial images using a game Arma3, enabling systematic analysis of view variation in aerial detection.
    We adopt the Arma3 \cite{arma3} rather than general-purpose game engines such as Unity \cite{unity} and Unreal \cite{unreal}.
    While those engines offer high-fidelity rendering, they often require custom asset modeling or significant licensing costs of domain-specific 3D models of equipment and terrain \cite{dosovitskiy2017carla, bala2024edify, yang2025genassets, blaga2025breaking}, for large-scale data construction.
In contrast, Arma3 already provides rich 3D object models with various terrain assets.
    This eliminates the need for heavy 3D modeling.

\subsection{Related work}
\noindent\textbf{Existing benchmark datasets for aerial object detection.}
        Early benchmarks, such as NWPU VHR-10 \cite{cheng2014multi}, DLR 3K Vehicle \cite{liu2015fast}, UCAS-AOD \cite{zhu2015orientation}, HRSC \cite{liu2016ship}, CARPK \cite{hsieh2017drone}, and RSOD \cite{long2017accurate}, primarily rely on static nadir-view imagery collected from Google Earth or airborne platforms.
        % \footnote{We summarize in Tab. \ref{tab1} the representative benchmarks for 2D object detection research in visible aerial imagery, acknowledging that not all existing datasets are included. 
    % While some datasets focusing on other tasks, such as aerial tracking (e.g., UAVDT \cite{du2018unmanned}, MOR-UAV \cite{mandal2020mor}) or scene graph generation (e.g., AeroEye series \cite{nguyen2024cyclo, nguyen2025thyme}), may also be utilized, we deliberately exclude them due to space limitations and to maintain a clear focus on datasets explicitly built for aerial object detection.}
    These datasets, typically consist of fewer than 2,000 images and limited to horizontal bounding boxes (HBB).
        They offer only coarse variations in viewpoint and scale, restricting their utility for robust generalization.
    Subsequent large-scale efforts, including DOTA-v2 \cite{ding2021object}, DIOR-R \cite{cheng2022anchor}, FAIR1M \cite{sun2022fair1m}, and DroneVehicle \cite{sun2022drone}, expand coverage with tens of thousands of images and a greater number of categories (up to 20) with oriented bounding boxes (OBB) to better capture object rotations.
        To further expand data diversity, Pan et al. \cite{pan2025locate} propose LAE-1M, which unifies ten existing remote sensing (RS) datasets, such as xView \cite{lam2018xview}, DOTA \cite{xia2018dota}, and DIOR \cite{li2020object}, into a single large-scale benchmark.
    Nevertheless, these datasets still largely center on single-view and nadir-oriented captures, lacking in multi-angle observations of the same scene (i.e. \textit{homogeneous aerial perspective}).

A few efforts to bridge multiple viewpoints have also emerged.
    For instance, MAVREC \cite{dutta2024multiview} is the first attempt to integrate aerial and ground viewpoints in a temporally synchronized manner, allowing simultaneous observations of the same scene from dual perspectives.
        Though datasets, like MOHR \cite{zhang2021empirical}, DroneVehicle \cite{sun2022drone}, and DVTOD \cite{song2024misaligned}, also provide multiple viewpoints, they do not ensure viewpoint simultaneity\footnote{Recently, the dataset MATIR-OD \cite{jiang2025object} has been collected and released using a DJI Matrice 300 drone-based camera to investigate how different observation angles affect aerial object detection in the infrared domain. While MATIR-OD provides valuable insights into multi-angle thermal detection, it focuses on a single category and a modest-scale collection (3,000 images with 43,540 instances) without simultaneous multi-view captures.}.
We argue that multi-perspective alignment in data represents an important step toward modeling cross-view understanding.

Hence, our AMOD dataset is proposed as a synthetically constructed, multi-angular visible aerial object detection benchmark based on Arma3 \cite{arma3}.
    While our dataset does not surpass recent large-scale datasets in terms of volume or category diversity, it offers a qualitatively distinct contribution: Unlike prior benchmarks focused on scale expansion, our dataset is captured from six distinct observation angles (i.e. \textit{simultaneous multiple viewpoints}), enabling systematic analysis of viewpoint variation on aerial detection.
% (36K images with 190K OBB annotations in 12 classes)

\noindent\textbf{Multi-view aerial image understanding.} Several lines of research in Earth vision leverage multi-view signals.
For example, Weir et al. \cite{weir2019spacenet} introduces a benchmark entitled MVOI to reveal how off-nadir imagery degrades building extraction performance, emphasizing the geometric and radiometric challenges that arise with varying observation angles. 
    The MVOI dataset provides multi-angle satellite imagery of the same scenes captured nearly simultaneously, but due to high annotation costs, only the nadir-view images are manually labeled; these annotations are then reused for off-nadir views, limiting the precision of view-specific supervision.
        In contrast, we generate our data using the Arma3 game, which allows us to provide far more precise bounding box labels for all viewing angles.
% MMGeo \cite{ji2025mmgeo}

Another important direction concerns cross-view geo-localization and matching \cite{ji2025mmgeo}.
    Ji et al. \cite{ji2025game4loc} introduces Game4Loc, a benchmark constructed using the GTA V \cite{gta5} environment to facilitate this research line.
        These studies demonstrate the potential of simultaneous multi-view data for geometric reasoning and spatial correspondence, though their primary focus lies in geo-localization and retrieval rather than category-level detection.
    In comparison, AMOD explores how viewpoint diversity influences category-level detection performance, providing a complementary perspective on multi-view aerial understanding.

\subsection{General setup} \vspace{-0.0cm}

\textbf{Category design.} As Arma3 is a military FPS game, most of its available 3D assets are related to military equipment.
    We adopt a total of 513 models in Arma3, 258 from the Eastern and 255 from the Western faction.
        Each model is then categorized into one of 12 classes: \textbf{Armored}, \textbf{Artillery}, \textbf{Helicopter}, \textbf{Landing Craft Utility (LCU)}, \textbf{Multiple Launch Rocket System (MLRS)}, \textbf{Plane}, \textbf{RADAR}, \textbf{Surface-to-air Missile (SAM)}, \textbf{Self-propelled Artillery}, \textbf{Support Vehicle}, \textbf{Tank}, and \textbf{Transporter Erector Launcher (TEL)}; the distribution of Arma3 assets assigned to each category is illustrated in \textcolor{purple}{Fig. 3 (Main)}.

\noindent\textbf{Objects placement (spawn).} Let $\mathcal{A}$ denote an area of interest (or target imaging area) for which simultaneous multi-view captures are provided as in \textcolor{purple}{Fig. 1 (Main)}.
    For each $\mathcal{A}$, our data generator\footnote{The values of $k$ and $e$ are empirically chosen and can be changed.} randomly selects and positions items as follows:
    \begin{itemize}
        \item \textbf{Step 1.} Sample $k$ classes among the above 12 categories, and store them into a set $S$. Here, $k$ is chosen between 1 and 8.
        \item \textbf{Step 2.} Randomly determine the number of items $e$ to be placed in the $\mathcal{A}$ ($e$ is between 8 and 14).
            The $e$ items are selected from the 513 assets, restricted to the classes in $S$.
    \end{itemize}

To further enhance data diversity, each object is assigned a unique rotation angle.
    Besides, we enforce that land assets must be located only on land and naval assets for sea, and that no objects overlap with each other.

\noindent\textbf{Recording setup.} As shown in Fig. \ref{amod:fig:teaser}, we control the Arma3 to capture six simultaneous multi-view, RGB-T paired images for each $\mathcal{A}$, corresponding to observation angles of 0$^\circ$, 10$^\circ$, 20$^\circ$, 30$^\circ$, 40$^\circ$, and 50$^\circ$.
    Along with the images, the associated metadata is also extracted to generate bounding-box annotations for each view.
Each image has a size of 1920$\times$1440 and the cameras are configured such that the central pixel of the 0$^\circ$-observing image is a ground sampling distance (GSD) of 0.1 m/pixel.
    For this nominal GSD, we set the camera altitude ($z$) to 120 m with Vertical FOV\footnote{VFOV 75$^\circ$ is a broad viewing angle similar to that of a standard wide-angle camera. As our primary focus is on unprocessed UAV imagery, neither orthorectification nor lens distortion correction is applied. Nevertheless, pretraining with AMOD improves performance over the baseline when evaluated on real-world datasets like DIOR-R, composed of satellite images from Google Earth (Standard Orthophoto).} 75$^\circ$ in Arma3.
        When $z$ exceeds 120 m, shadows largely disappear, resulting in a notable loss of image detail in Arma3.
            Since the presence of shadows in aerial imagery is known to be of significant research importance \cite{liu2023shadow}, we do not consider $z>$120 m.
                Besides, we set the time range from 9 to 18 for capturing different lighting conditions\footnote{In Arma3, visible images captured at night are usually rendered entirely black, making them unusable. Therefore, we do not collect RGB-T pairs under nighttime conditions.}.
Meanwhile, to enlarge the geographical coverage, we leverage several official Arma3 maps as follows\footnote{Note that all maps are utilized at a nearly equal frequency.}: 
\begin{itemize}\small
    \item For train/validation splits: Altis, Malden, Stratis, Tanoa, Weferlingen;
    % , and Weferlingen (Winter);
    \item For test split: Malden and Livonia.
\end{itemize}

\noindent\textbf{Annotation.} While capturing each scene, our generator obtains the 3D bounding box using `boundingBoxReal' function of Arma3; \textcolor{blue}{the detailed pipeline is presented in Sec. A (Supp. Material)}. It then projects its corner points onto the image plane, and derive both horizontal and oriented bounding box (HBB/OBB) annotations.
    The HBB is computed by taking the minimum and maximum projected coordinates, while the OBB is obtained by applying a convex hull \cite{graham1972efficient} followed by rotating calipers \cite{toussaint1983solving} to find the minimum-area enclosing rectangle.
% However, OBBs obtained via the rotating calipers algorithm can be suboptimal (i.e., relatively loose) at certain instances. 
%         Since tighter bounding boxes are generally more desirable for detector training and also beneficial for fair benchmarking \cite{rezatofighi2019generalized, murrugarra2022can, xu2022detecting}, we further refine the initial OBBs. 
%     Specifically, we employ the Segment Anything Model (SAM) \cite{kirillov2023segment} to extract object masks from the initial OBBs, and subsequently reconstruct tighter OBBs based on these masks. 
%     \textcolor{purple}{Detailed refinement procedure and its relevant source code are provided in Supp. Material and our online website.}
Finally, as summarized in \textcolor{purple}{Tab. 2 (Main)}, in our benchmark, to facilitate future multi-view studies, the same object observed across different views is assigned a consistent object ID for each region of interest ($\mathcal{A}$).

\noindent\textbf{Data balancing.} 
Class imbalance is undesirable, as it may bias model training toward overrepresented categories and degrade generalization performance \cite{kim2021imbalanced}.
    However, achieving perfectly uniform class distributions is infeasible in our data because the number of available 3D assets varies across categories as seen in \textcolor{purple}{Fig. 3(a) (Main)}.
To address this, we post-process our initially generated data so that the ratio between the number of instances per class and the number of Arma3 items is made as uniform as possible; see \textcolor{purple}{Fig. 3(c) (Main)}.

\vspace{-0.1cm}
\subsection{Statistics, challenges, and applications} 

\textbf{Train, validation, and test splits.} The dataset is partitioned into three subsets: a training set comprising 47,808 images (64.68\%), a validation set with 11,988 images (16.21\%), and a test set with 14,124 images (19.11\%).
    \textcolor{black}{We maintain similar class distributions across all splits as shown in \textcolor{purple}{Fig. 3(b-2) (Main)}.}

\noindent\textbf{Object size and aspect ratio distributions.} 
Following \cite{dutta2024multiview}, we categorize object sizes into three groups: small ($\le$ 32$\times$32 pixels), medium (between 32$\times$32 and 96$\times$96), and large ($\ge$ 96$\times$96). The object size distribution of our dataset is illustrated in \textcolor{purple}{Fig. 3(d) (Main)}.
    Meanwhile, \textcolor{purple}{Figs. 3(e)-(f) (Main)} illustrate the average size and aspect ratio of OBBs for each category, respectively.

\begin{figure}[!ht]
\centering
\vspace{-0.0cm}
\includegraphics[width=6cm]{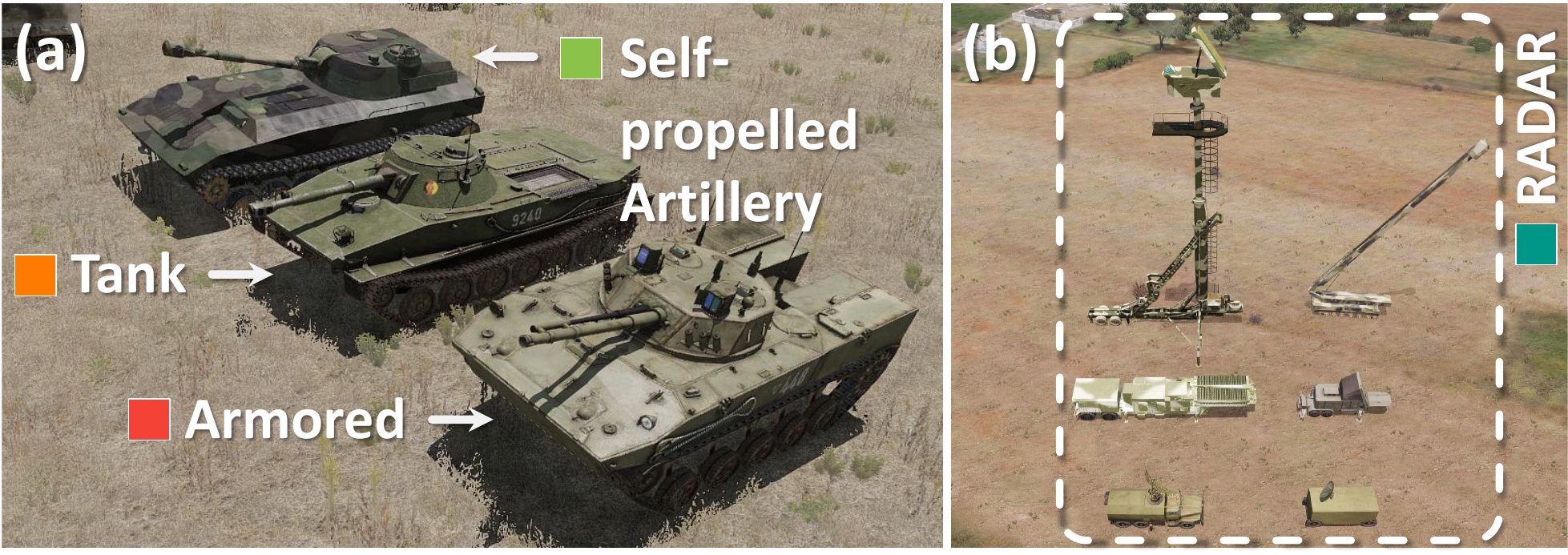}
  \vspace{-0.2cm}
  \caption{Examples of (a) low inter-class and (b) high intra-class variances in Arma3 models we used for our dataset.} \vspace{-0.2cm}
  % \Description{Enjoying the baseball game from the third-base
  % seats. Ichiro Suzuki preparing to bat.}
  \label{amod:fig:4}
\end{figure}

\noindent\textbf{Distinct characteristics of AMOD.} Our dataset is designed with three notable properties as follows.
\begin{enumerate}
    \item It provides \textit{simultaneous multi-view observations} of each region of interest ($\mathcal{A}$), enabling research on cross-view consistency in aerial image understanding.
        Owing to the complex terrain in Arma3 environments, such as cliffs and mountains, certain objects are occluded in some viewpoints\footnote{We automatically filter objects as fully occluded if their direct line-of-sight intersects with other scene elements.}.
    \item It exhibits both \textit{low inter-class and high intra-class variations}.
    For instance, visually similar vehicle types such as tanks and armored vehicles challenge fine-grained recognition, while a single class like RADAR includes highly diverse structures, as illustrated in Fig. \ref{amod:fig:4}.
    \item It ensures \textit{diverse geographical backgrounds} by leveraging diverse official Arma3 maps, thereby enhancing scene variability and model's generalization ability.
\end{enumerate}

\begin{table}[!t]
\vspace{-0.0cm}
\centering
\caption{Class-wise Test AP on AMOD.} \vspace{-0.4cm}
\label{amod:tab:2}
\resizebox{8.5cm}{!}{%
\begin{tabular}{@{}lccccccccccccc@{}}
\toprule
 & Armored & Artillery & Helicopter & LCU & MLRS & Plane & RADAR & SAM &
\begin{tabular}[c]{@{}c@{}}Self-propelled\\Artillery\end{tabular}
& Support & Tank & TEL & Mean \\ \midrule
AP$_{75}$ & 40.30 & 74.20 & 50.40 & 78.30 & 65.20 & 79.30 & 25.70 & 30.60 & 37.40 & 40.60 & 47.40 & 53.70 & 51.93 \\
AP$_{50:95}$ & 47.19 & 64.35 & 50.47 & 58.27 & 56.08 & 65.20 & 31.65 & 38.92 & 43.53 & 46.46 & 48.91 & 54.76 & 50.48 \\ \bottomrule
\end{tabular}
} \vspace{-0.cm}
\end{table}

\begin{figure}[!t]
    \centering
    \includegraphics[width=8.5cm]{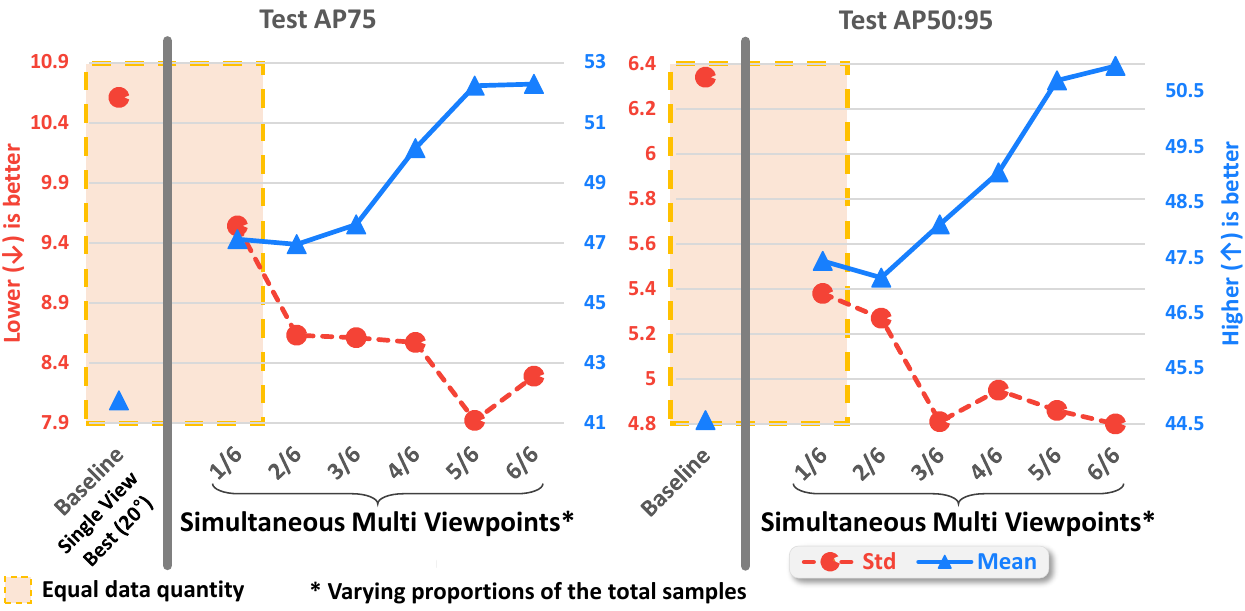} \vspace{-0.1cm}
    \caption{Effect of angular diversity versus data quantity on AMOD (AP$_{75}$, AP$_{50:95}$). The \textcolor{yechanblue}{blue} and \textcolor{red}{red} dots denote the mean and standard deviation of AP values across all observation angles, respectively.}
    % Increasing data quantity is dominant for performance, but under equal data quantity (\textcolor{yechanmandarin}{\textbf{$\square$}}), angular diversity leads to higher mAP and lower variance across angles.
    \label{amod:fig:5}
    \vspace{-0.cm}
\end{figure}

\subsection{Detailed experimental setup and more experimental results for our AMOD benchmark dataset}
\label{multiviewsup}

Unless otherwise stated, all experiments here are conducted with the same detector, dataset, and optimization configuration.
We adopt Oriented R-CNN with a Swin-S backbone as implemented in MMRotate, initialized from ImageNet-pretrained weights.

% \vspace{+0.2em}
% \noindent\textbf{Dataset and angle splits.}
% For all experiments here, we use the proposed AMOD dataset (RGB split only) with a default image resolution of $1920\times1440$ pixels.

\vspace{+0.2em}
\noindent\textbf{Class-wise performance on AMOD} \label{class-wise}
We first report class-wise detection performance of AMOD in Tab.~\ref{amod:tab:2}, which serves as a reference breakdown of category-level performance under the default evaluation setting. Categories such as LCU (78.30 AP$_{75}$) and Plane (79.30 AP$_{75}$) achieve relatively high performance, likely because they exhibit distinctive global shapes with limited visual ambiguity. In contrast, categories such as RADAR (25.70 AP$_{75}$) and SAM (30.60 AP$_{75}$) show substantially lower performance. We hypothesize that this is partly attributable to high intra-class variation; as shown in Fig.~\ref{amod:fig:4}(b).

\vspace{+0.2em}
\noindent\textbf{Multi-angle training.} As shown in \textcolor{purple}{Tab. 3 (Main)}, the model trained with all view angles (``All'') achieves the highest overall performance compared to single-angle training.  
    % This result demonstrates the clear benefit of using simultaneous multi-viewpoints, which expose the detector to diverse geometric contexts.
However, one may argue that this improvement simply stems from the increased amount of training data, since each single-angle model only sees one-sixth of the total samples.
    To disentangle the effect of angular diversity from that of data quantity, we conduct an additional experiment as illustrated in Fig. \ref{amod:fig:5}.
\begin{itemize}
    \item \textbf{\textit{Data quantity (or scene diversity) is the dominant factor for overall performance gain}}. As the total data quantity decreases, the overall test mAP gradually drops, while the AP variance across look angles increases.
    % , indicating reduced stability and generalization.
    \item \textbf{\textit{Yet, when controlling for data quantity, we find that angular diversity itself plays a crucial role in enhancing generalization}}, leading to both higher mAP and lower performance variance across look angles.
\end{itemize}

% \noindent\textbf{Dataset and angle splits.}
%     For \textit{single-angle} training, we restrict the training set to one of these angles and keep the validation/test sets unchanged (i.e. all angles are used individually).
% For \textit{multi-angle} training, all six angles are used jointly during training.
% For the reduced-data experiments in Fig. \ref{amod:fig:5}, we further subsample training scenarios with a nested strategy (e.g., $3/6 \subset 4/6 \subset 5/6 \subset 6/6$) while always preserving the full set of angles.

\begin{figure*}[!t]
    \centering
    \includegraphics[width=15cm]{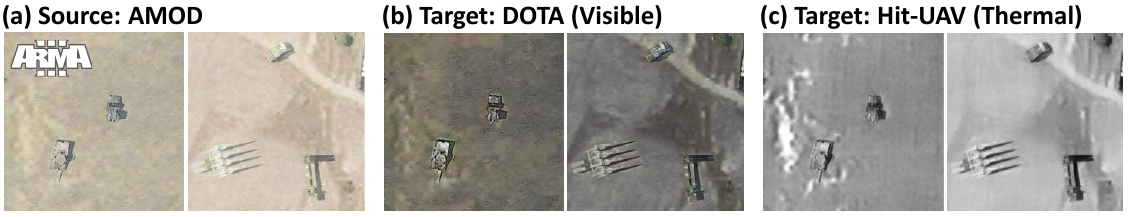} \vspace{-0.1cm}
    \caption{Examples of translated synthetic image from AMOD using UNIT.}
    \label{supfig7}
    \vspace{-0.2cm}
\end{figure*}

\section{Real-world Domain Adaptation}
% \section{Domain Transfer, Pretraining, and\\\hspace*{-0.5em}Finetuning for Real-world Generalization}

\subsection{Preliminary Background}
Though our synthetic data offers scalable and precisely annotated samples, the visual characteristics of synthetic images significantly differ from real-world data in aspects such as color tone, texture, and background complexity \cite{wang2021pixel}.
    This discrepancy is generally referred to as \textit{domain gap}.
Even in real-world scenarios, domain gaps can still naturally arise due to varying environmental factors, sensors, and imaging modalities \cite{wu2017rgb, baek2024unexplored}. % wang2019learning
    % Consequently, effectively transferring features between such heterogeneous domains becomes a crucial challenge, known as \textit{domain adaptation}.
    Note that the thermal version of our AMOD utilizes the Arma 3 TI engine to generate white-hot thermal imagery. If a specific spectrum like MWIR is required, users may convert the output using image-to-image translation methods.

\subsection{Overall pipeline}
\label{domain_adaptation}
\begin{figure}[!t]
    \centering
    \includegraphics[width=8.3cm]{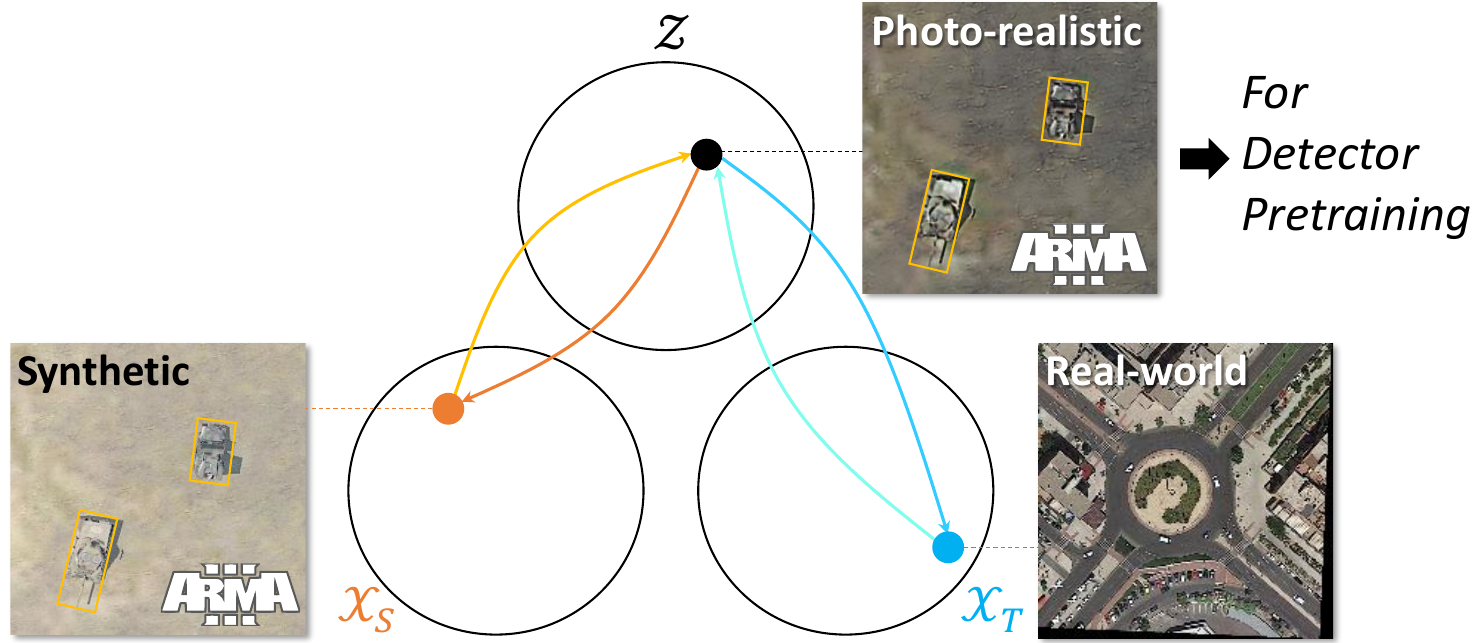} \vspace{-0.1cm}
    \caption{Illustration of the shared latent space ($\mathcal{Z}$) for domain adaptation between our synthetic ($\mathcal{X}_{S}$) and target real-world ($\mathcal{X}_{T}$) aerial image domains, inspired by UNIT \cite{liu2017unsupervised}. Both domains are mapped to $\mathcal{Z}$, from which Arma3-rendered photo-realistic images are generated as intermediate samples to bridge $\mathcal{X}_{S}$ and $\mathcal{X}_{T}$; see Fig. \ref{supfig7} for translated synthetic image examples (from $\mathcal{Z}$) of AMOD.}
    \label{supfig9}
    \vspace{-0.1cm}
\end{figure}

To bridge this gap, we adopt a pipeline consisting of domain transfer, pretraining, and finetuning a detection model, as shown in Fig. \ref{supfig11}.
    We employ UNIT \cite{liu2017unsupervised} to transform our synthetic AMOD images ($\mathcal{X}_{S}$) into photo-realistic ones that resemble the target domain ($\mathcal{X}_{T}$).
    % \footnote{Other generative models for unpaired image-to-image (i2i) translation, including CycleGAN \cite{zhu2017unpaired}, can also be adopted instead of UNIT.}
        Both domains $\mathcal{X}_{S}$ and $\mathcal{X}_{T}$ are embedded into a shared latent space ($\mathcal{Z}$), from which visually realistic and target-oriented images are augmented\footnote{The embedding in $\mathcal{Z}$ is fed into the generator $G_{S \rightarrow T}$ which transfers it from $\mathcal{X}_{S}$ to $\mathcal{X}_{T}$, producing a synthetic image in the target style. Both $G_{S \rightarrow T}$ and $G_{T \rightarrow S}$ are trained, yet we only utilize $G_{S \rightarrow T}$ of the UNIT.}, as illustrated in Figs. \ref{supfig7} and \ref{supfig9}.

After translating AMOD images, we pretrain the object detector on these samples to learn domain-invariant features that generalize well across synthetic and target real data.
    Subsequently, we finetune the pretrained detector using labeled real-world data to adapt to the final target distribution.
% \footnote{The class taxonomy of AMOD used during pretraining may not be perfectly aligned with that of the dataset used for finetuning. In such case, only the backbone is utilized during finetuning.}

% , while the remaining head layers are replaced and re-initialized with random weights.

\vspace{-0.2cm}
\subsection{Stochastic Boundary Smoothing (SBS): a thermal-specific augmentation trick for AMOD-pretraining
} \vspace{-0.0cm}
\label{pretraining_trick}
% \textbf{Motivation.} 
% Building upon Sec. \ref{domain_adaptation}, we explore not only the synthetic-to-real transferability within the same spectrum, but also visible-to-thermal cross-modal adaptation\footnote{It is important as thermal-infrared data are more expensive to collect than visible imagery \cite{ma2025self} and their quantity is often limited.} with the AMOD dataset.
    % For this, visible-to-thermal translation should be performed before pretraining a detector with our dataset.

\begin{figure}[!t]
    \centering
    \includegraphics[width=8.1cm]{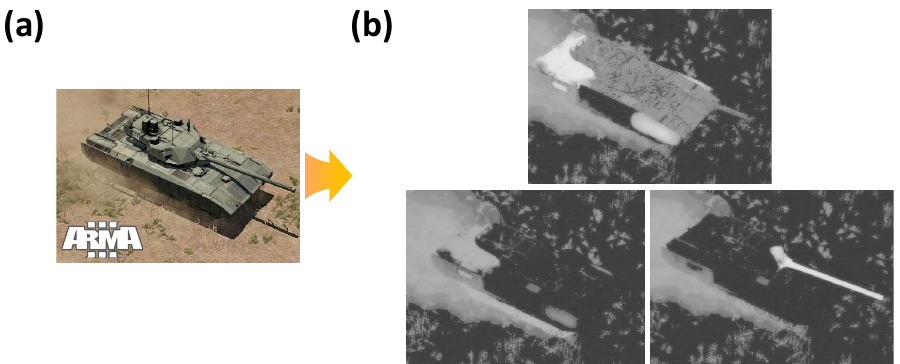} \vspace{-0.1cm}
    \caption{Example of visible-to-thermal variations. (a) shows the same visible image, while (b) presents multiple thermal renderings produced under varying ambient and object heat conditions in Arma3, with the time condition kept unchanged. It illustrates that thermal-modality translation is inherently ill-posed, as an identical optical input can correspond to diverse thermal outcomes.}
    \label{supfig10}
    \vspace{-0.4cm}
\end{figure}

As illustrated in Fig. \ref{supfig10}, the process of translation into thermal-spectrum is inherently \textit{ill-posed}.
    Consequently, this process is fundamentally a one-to-many translation, which no deterministic model can fully capture.
For instance, though UNIT \cite{liu2017unsupervised} we adopt produces visually coherent outputs, its deterministic nature results in limited representations that cannot reflect the diverse thermal variations encountered in real-world sensing.
    Unfortunately, existing studies \cite{ozkanouglu2022infragan, lee2023edge, han2024dr, li2025cross, zhao2025diverse, ma2025self, banday2025multi} largely treat such translation as a deterministic mapping.
    A straightforward remedy might be stochastic image translation \cite{zhao2022egsde, li2023bbdm, kim2023unpaired, lan2025schrodinger}.
        However, this direction often introduces cumbersome network architectures or complicated training procedures like multi-step sampling.

To address this challenge, we avoid modifying the translator itself and introduce stochasticity through augmentation instead.
    Specifically, we apply perturbations to the pseudo-thermal translated images during detector pretraining.
This allows us to rely on simpler deterministic translators while still expanding thermal variability.

To design a meaningful perturbation strategy, we take into account that thermal sensors suppress inner textures while preserving sharp boundary gradients, yet those boundaries can still become blurred due to weak thermal contrast or sensor smoothing \cite{hwang2015multispectral, cai2024exploring}.
    Motivated by this fact, we introduce \textbf{\textit{stochastic boundary smoothing}} (SBS), a simple but effective augmentation strategy during AMOD-pretraining for thermal-infrared detection, applied to each object at each iteration.
        Consequently, selectively smoothing contour mitigates the overly crisp edges in synthetic sources, effectively shrinking the domain gap, while improving thermal variability.

Let $f$ denote the transform applied to the image in each iteration during pretraining.
    In Fig. \ref{supfig11}, $f$ is basically an identity mapping.
However, for SBS, we replace $f$ with:
\vspace{-0.2cm}
\begin{equation}\label{eq1} \small
\setlength{\belowdisplayskip}{2pt}
\begin{split}
    f(\mathbf{X}) =\begin{cases}
    \mathbf{X}, & z = 0, \\
    \mathbf{X} \odot (1 - {\mathbf{C}}) + {G}_{\sigma}(\mathbf{X}) \odot {\mathbf{C}}, & z = 1,
    \end{cases}
     % f(x; \sigma) = x \odot (1-{m}_{b}) + (x*{k}_{\sigma}) \odot {m}_{b}, \\
\end{split} 
\vspace{-0.5cm}
\end{equation} 
where $\mathbf{X}$ is a translated image (size: $w\times h$) and $z$ is drawn from a Bernoulli distribution with a probability of $p$. 
    $\odot$ denotes the Hadamard product.
        ${G}_{\sigma}$ represents a 2D Gaussian blur operator parameterized by the standard deviation $\sigma$.
    $\mathbf{C}$ is the aggregated mask of contours over all instances as: \vspace{-0.6cm}

\begin{equation}\label{eq2} \small
\setlength{\belowdisplayskip}{2pt}
\begin{split}
    \mathbf{C} = {\bigvee}_{i} \,\, \Gamma(\mathbf{M}_i; \tau_i), 
    % \quad \tau_i \sim \mathcal{U}(\tau_{\min}, \tau_{\max}),
     % f(x; \sigma) = x \odot (1-{m}_{b}) + (x*{k}_{\sigma}) \odot {m}_{b}, \\
\end{split} 
\vspace{-0.0cm}
\end{equation} 
where  $\bigvee$ denotes a pixel-wise logical OR operator applied over all contour masks. $\mathbf{M}_i \in \{0, 1\}^{w\times h}$ is a mask of $i$-th object in $\mathbf{X}$. For $\mathbf{M}_i$, $\tau_i$ is a contour thickness randomly drawn from a uniform distribution $\mathcal{U}(\tau_{\min}, \tau_{\max})$. $\Gamma$ is a contour-band operator that produces a binary band (shape: $\{0, 1\}^{w\times h}$) that lies in the boundary of $\mathbf{M}_i$ with $\tau_i$ as: \vspace{-0.6cm}

\begin{equation}\label{eq3}  \small
\setlength{\belowdisplayskip}{1pt}
\begin{split}
\Gamma(\mathbf{M}_i;\tau_i)(x)
= \mathbf{1}\!\left[
\bigl\lvert \mathrm{dist}(x,\partial\mathbf{M}_i)\bigr\rvert \le \tau_i
\right], 
% \quad \forall_x, 
     % f(x; \sigma) = x \odot (1-{m}_{b}) + (x*{k}_{\sigma}) \odot {m}_{b}, \\
\end{split} 
\end{equation} 
where $x$ denotes each pixel of $\mathbf{X}$ and $\mathrm{dist}(x,\partial\mathbf{M}_i)=\mathrm{min}_{y \in \partial\mathbf{M}_i} || x-y ||_2$. $\partial\mathbf{M}_i$ is a contour of $i$-th object. 

% \begin{equation}\label{eq4} 
% \begin{split}
% \partial M_i
% = \mathbf{1}\!\left[
% \,\mathbf{M}_i(x)=1 \ \land\ 
% \exists\, y\in\mathcal{N}(x):\ M_i(y)=0\,
% \right]_x.
%      % f(x; \sigma) = x \odot (1-{m}_{b}) + (x*{k}_{\sigma}) \odot {m}_{b}, \\
% \end{split} 
% \vspace{-0.5cm}
% \end{equation} 

\begin{table}[t!]
\vspace{-0cm}
\centering
\caption{Evaluation results (AP$_{50}$) on HIT-UAV. Effect of the SBS perturbation ($f$: SBS) under AMOD pretraining.}\resizebox{7cm}{!}{%
\begin{tabular}{@{}ll|ccccc@{}}
\toprule
\multicolumn{2}{c|}{Pretrained from} & \multicolumn{5}{c}{Finetuned for HIT-UAV} \\
\cmidrule(lr){1-2}\cmidrule(l){3-7}
Source Data & Pretraining-specific perturbation
& Person & Car & Others & Mean & $\Delta$ \\
\midrule

ImageNet & --
& 85.30 & 81.29 & 57.12 & 74.57 & -- \\
\midrule

\multirow{2}{*}{AMOD}
& \xmark
& 86.59 & 82.00 & 58.51 & 75.70 & \textcolor{red}{+1.13} \\
& \cmark\ ($f$: SBS)
& 87.68 & 82.15 & 61.40 & 77.08 & \textcolor{red}{+2.51} \\
\bottomrule
\end{tabular}%
} \vspace{-0.3cm}
\label{sbshituav}
\end{table}

\begin{figure}[!t]
    \centering
    \includegraphics[width=8cm]{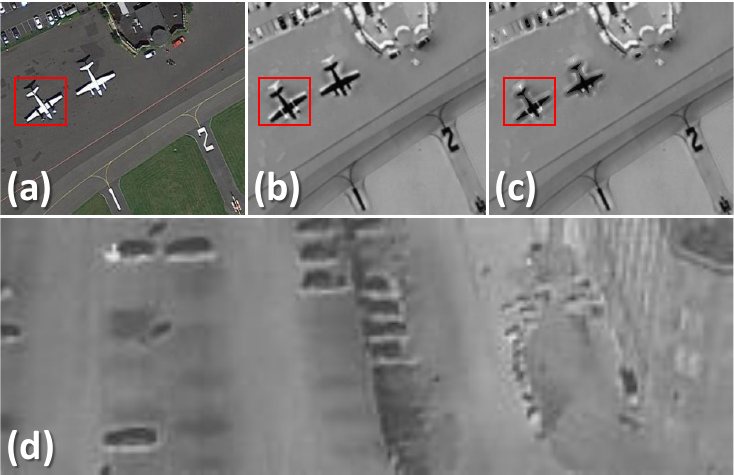} \vspace{-0.1cm}
    \caption{(a) Example visible image from DIOR-R.
(b) thermal-style translation of (a) using a pretrained UNIT model.
(c) Result after applying the proposed Stochastic Boundary Smoothing (SBS), where the object boundaries become intentionally blurred to better emulate the characteristic edge attenuation observed in real airborne thermal sensors.
(d) Example real thermal image from HIT-UAV.}
    \label{supfig8}
    \vspace{-0.2cm}
\end{figure}

For SBS, each instance-wise mask $\mathbf{M}_i$ is automatically extracted by Segment Anything Model (SAM) \cite{kirillov2023segment} with bounding box prompts.
% ; see Sec. \ref{SAMBA} for implementation details of SBS. 
    Note that while SBS is designed for AMOD-based pretraining, it could potentially be extended to other datasets as well, as shown in Fig. \ref{supfig8}.
        SBS can be seamlessly incorporated into other datasets as well, as can be seen in Fig. \ref{supfig8}.
We begin by extracting object masks from a DIOR-R visible (see Fig. \ref{supfig8}(a)) image using SAM, translate the image into a thermal-like domain with a pretrained UNIT model (see Fig. \ref{supfig8}(b)), and then apply SBS (see Fig. \ref{supfig8}(c)).
    The difference between (b) and (c) shows that SBS deliberately smooths object boundaries to replicate the characteristic edge attenuation observed in real airborne thermal sensors (see Fig. \ref{supfig8}(d)).
This observation confirms that SBS can serve as a general-purpose augmentation module by boundary smoothing for enhancing thermal-style realism across diverse datasets.
        We defer this to future work, as the primary focus of this paper is on the AMOD dataset.

Adopting our \textbf{\textit{stochastic boundary smoothing}} (SBS) (see Sec. \ref{pretraining_trick}) during pretraining consistently leads to additional performance gains. For all experiments, $p=0.5$, $\sigma=11$, $\tau_{\min}=2$, $\tau_{\max}=5$ are used in SBS.
    The performance reported on the HIT-UAV thermal benchmark in \textcolor{purple}{Tab. 4 (Main)} is obtained after applying AMOD pretraining, where our SBS method is incorporated during the pretraining stage.
        A comparison of performance before and after applying SBS is provided in Tab. \ref{sbshituav}.
    % \textbf{\textit{This simple perturbation helps narrow the thermal gap by alleviating overly sharp boundaries of source visible images, while mimicking the spatial patterns of thermal imagery}}.

\vspace{-0.2cm}
\subsection{Detailed experimental setup for real-world generalization}
\label{setting53}

\begin{figure}[!t]
    \centering
    \includegraphics[width=8.1cm]{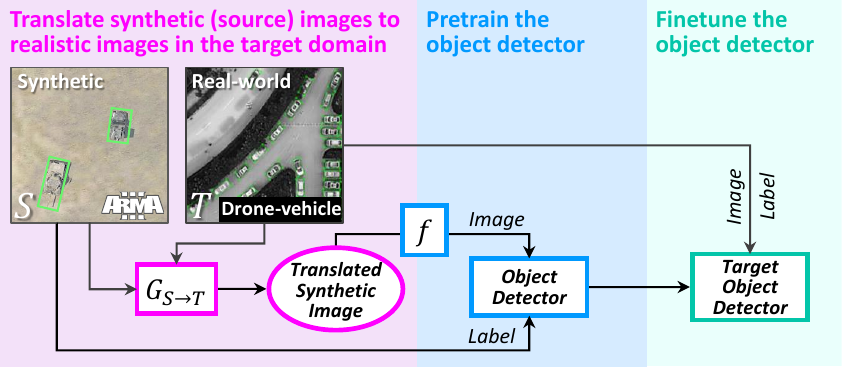} \vspace{-0.1cm}
    \caption{Overall domain adaptation pipeline used in this work. Here, $\mathcal{X}_{T}$ is a proxy target-style domain used for translation during pretraining, which may differ from the final downstream benchmark used for finetuning/evaluation.}
    \label{supfig11}
    \vspace{-0.1cm}
\end{figure}

\noindent\textbf{Pretraining setup.} We follow the same training configuration described in Sec. \ref{multiviewsup}, except for the training epochs.
    The detector is pretrained for 5 epochs.  
Before pretraining, AMOD is translated into the corresponding target-domain style.

\vspace{+0.2em}
\noindent\textbf{Finetuning setup.}
After pretraining, only the backbone weights are transferred to the downstream detector.  
    The detection head is \textbf{re-initialized} to match the number of classes of the target dataset (e.g., DIOR-R, DroneVehicle, HIT-UAV).  
Finetuning is conducted for 25 epochs, with all optimizer parameters (learning rate, momentum, weight decay, LR schedule, and warm-up) based on the conventional training protocol used for each target dataset.

\end{document}